%% file: main.tex
\documentclass[10pt,twocolumn,letterpaper]{article}

\usepackage{iccv}              


\definecolor{iccvblue}{rgb}{0.21,0.49,0.74}
\usepackage[pagebackref,breaklinks,colorlinks,allcolors=iccvblue]{hyperref}

\def\paperID{4} 
\def\confName{ICCV}
\def\confYear{2025}

\title{FLUXSynID: A Framework for Identity-Controlled Synthetic Face Generation with Document and Live Images}

\author{
  Raul Ismayilov \quad Dzemila Sero \quad Luuk Spreeuwers \\
  University of Twente \\
  Enschede, Netherlands \\
  {\tt\small \{raul.ismayilov, d.sero, l.j.spreeuwers\}@utwente.nl}
}

\begin{document}
\maketitle

\begin{abstract}
    Synthetic face datasets are increasingly used to overcome the limitations of real-world biometric data, including privacy concerns, demographic imbalance, and high collection costs. However, many existing methods lack fine-grained control over identity attributes and fail to produce paired, identity-consistent images under structured capture conditions. We introduce FLUXSynID, a framework for generating high-resolution synthetic face datasets along with a dataset of 14,889 synthetic identities. We generate synthetic faces with user-defined identity attribute distributions, offering both document-style and trusted live capture images. The dataset generated using the FLUXSynID framework shows improved alignment with real-world identity distributions and greater inter-class diversity compared to prior work. Our work is publicly released\footnote{\url{https://github.com/Raul2718/FLUXSynID}} to support biometric research, including face recognition and morphing attack detection.
\end{abstract}

\section{Introduction}
Large-scale, diverse, and well-annotated face datasets are crucial for training and evaluating biometric systems. In addition to face recognition~\cite{Melzi2024FRCSynonGoingBA, 10216308, Robbins2024DaliIDDL} and emotion analysis~\cite{Roy2024ResEmoteNetBA, Elsheikh2024ImprovedFE}, recent work on Morphing Attack Detection (MAD)~\cite{Chaudhary2021DifferentialMF, Liu2024DifferentialMA, Banerjee2021ConditionalID} and face demorphing~\cite{10415238, Shukla2024FacialDV} relies significantly on such datasets. However, collecting real-world facial data presents substantial challenges. Privacy regulations restrict access and usage, while the high financial cost of acquiring and annotating large volumes of data can be prohibitive. Moreover, demographic imbalances and the need for operationally relevant conditions, e.g., capturing both document-style and trusted live capture images for the same subject, further complicate data collection efforts.

In particular, applications such as Differential Image-Based MAD (D-MAD)~\cite{face_morphing_attacks} require datasets containing paired, identity-consistent document-style and live capture images. In typical scenarios, such as border control, the document image may be a bona fide or morphed photo, while the live capture serves as a trusted reference. These tasks also require demographically diverse identities to ensure fairness and robustness. Yet, existing datasets are often limited in scale, and collecting real data with such properties raises legal, ethical, and logistical challenges.

While recent advances in generative models have facilitated the creation of synthetic datasets~\cite{gandiffface, Boutros2022SFacePA, Qiu2021SynFaceFR, Boutros2022UnsupervisedFR, stylegan2}, most focus on in-the-wild imagery and lack support for structured capture conditions or fine-grained control over identity attributes (e.g., age, ethnicity). Moreover, generated faces often diverge from the distributional characteristics of real data~\cite{Huber2023BiasAD}. The ONOT dataset~\cite{onot} addresses some of these issues but remains limited in scale, control, and inter-class diversity.

In this work, we introduce FLUXSynID, a framework for generating synthetic face datasets with controllable identity attributes and paired document-style and live capture images. Built on the FLUX.1 [dev]~\cite{flux} diffusion model and enhanced via LoRA-based fine-tuning~\cite{lora}, FLUXSynID produces high-quality, identity-consistent images guided by natural language prompts. It employs a two-stage pipeline: generating document-style images first, then corresponding live capture variants using three complementary methods. These are based on LivePortrait~\cite{live_portrait}, Arc2Face~\cite{arc2face}, and PuLID~\cite{pulid}, each introducing different levels of variation in pose, expression, lighting, and background.

The contributions of this work are as follows:
\begin{itemize}
  \item A method for generating high-quality document-style face images with user-defined identity attribute distributions, enabling direct demographic control.
  \item A novel live image generation approach that adapts the FLUX.1 [dev] and PuLID frameworks with new conditioning strategies for identity and structure preservation. Two complementary methods, LivePortrait and Arc2Face, are also integrated to generate multiple, diverse face images per identity, supporting downstream tasks such as D-MAD.
  \item A comparative evaluation against real and synthetic face datasets, demonstrating closer alignment of FLUXSynID with real data in both visual appearance and embedding-space distributions. The demographic bias of the generative model is also investigated and reported.
  \item The public release of the FLUXSynID framework and a dataset of 14,889 synthetic identities to support biometric research.
\end{itemize}

\section{Related Work}
Synthetic face datasets are increasingly used in biometric research due to the limitations of real facial data, such as privacy concerns, high costs, and demographic imbalances~\cite{Huber2023BiasAD, gandiffface, onot, Boutros2022SFacePA}. SynFace~\cite{Qiu2021SynFaceFR} and DigiFace-1M~\cite{Bae2022DigiFace1M1M} showed promise but offered limited variation, realism, and identity control, restricting their applicability to tasks such as Morphing Attack Detection (MAD)~\cite{face_morphing_attacks}.

GANDiffFace~\cite{gandiffface} combines StyleGAN3~\cite{Karras2021AliasFreeGA} with linear SVMs to control demographic attributes in synthetic face generation, and introduces intra-class variation via per-identity diffusion model fine-tuning~\cite{dreambooth}. While this allows for more controlled sampling, the method is computationally expensive and requires a separate SVM for each attribute. GANDiffFace, similar to SFace~\cite{Boutros2022SFacePA}, primarily targets unconstrained images and lacks support for document-style outputs. Despite offering attribute control, GANDiffFace lacks scalability and alignment with structured document-style generation needs.

Recent studies have leveraged diffusion models for identity-consistent face generation, achieving high realism and intra-class variation. DCFace~\cite{Kim2023DCFaceSF} combines identity and style embeddings to synthesize diverse images of the same person, while IDiff-Face~\cite{Boutros2023IDiffFaceSF} conditions a latent diffusion model on feature representations, achieving similar identity consistency. While effective, both methods focus on unconstrained synthesis and offer limited control over document-style formatting or demographic attributes.

The ONOT dataset~\cite{onot} represents a pioneering effort in document-style face synthesis for biometric use, employing a fine-tuned Stable Diffusion~\cite{latent_diffusion} model and prompt engineering to generate ICAO-compliant~\cite{icao} ID images. It sampled 64 images per identity across 15k identities, retaining only those with at least one compliant image, yielding usable samples for just 27\% of identities. Further intra- and inter-class filtering reduced the final dataset to 255 identities. Our analysis shows that retained identities were highly similar and poorly aligned with the real identity embedding space. ONOT's document-style control relies solely on prompts, lacking the fine-tuning needed for consistent formatting. Its demographic attribute control is rigid and not easily extensible. While ONOT shares our high-level goals, it falls short in flexibility, scalability, and diversity.

\section{FLUXSynID: Document Images} \label{sec:doc_imgs}
\begin{figure*}
\centering
\includegraphics[width=\linewidth]{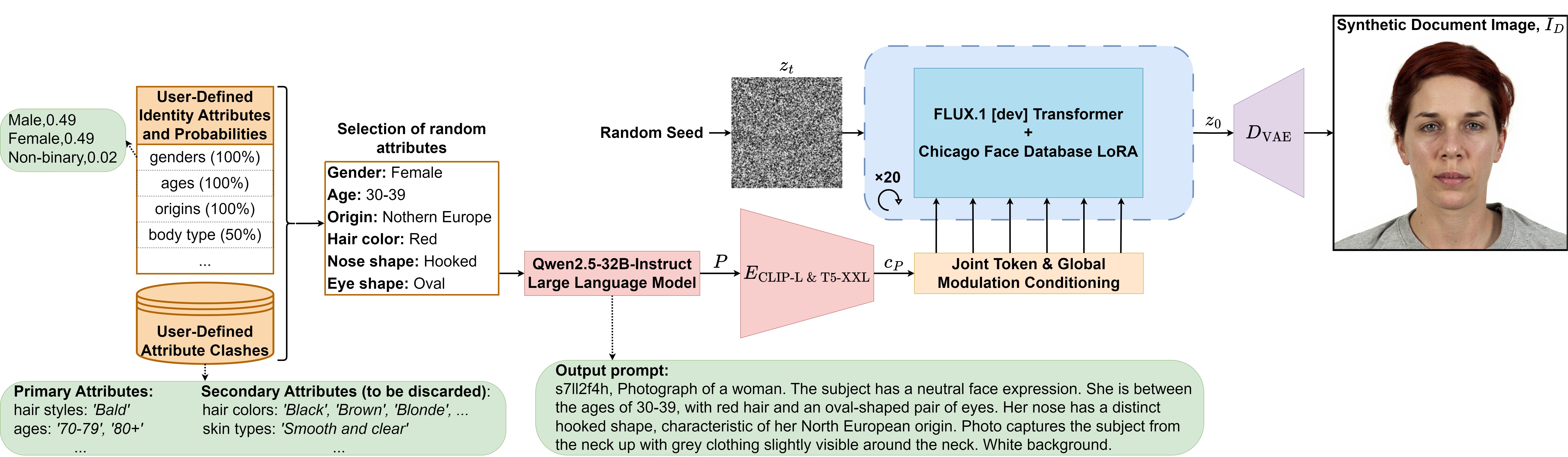}
\caption{Overview of the FLUXSynID pipeline for synthetic document image generation. Given user-defined identity attributes and constraints, descriptive prompts are generated and used to guide FLUX.1 [dev]~\cite{flux} diffusion model enhanced with LoRA adapter~\cite{lora} for producing realistic frontal document-style images. Please refer to \cref{sec:doc_imgs} for details.}
\label{fig:document_images_workflow}
\end{figure*}

Frontal, standardized face images, such as those found in passports, are essential for biometric tasks such as MAD~\cite{face_morphing_attacks}. Unlike unconstrained imagery, these images must meet strict format standards: frontal pose, neutral expression, consistent lighting, and plain backgrounds. Generating such images synthetically is challenging because most diffusion or GAN-based models are trained on in-the-wild datasets. Moreover, existing methods often lack the control needed to ensure demographic diversity, leading to biased or unrealistic datasets.

\Cref{fig:document_images_workflow} presents an overview of the FLUXSynID pipeline for synthetic document image generation, which centers around the FLUX.1 [dev]~\cite{flux} diffusion model. The process begins with user-defined identity attributes, which are used to generate natural language prompts via a large language model. These prompts guide FLUX.1 [dev], fine-tuned with a LoRA adapter~\cite{lora}, to produce realistic, high-resolution document-style images.

\subsection{Identity Attributes}
Existing face image datasets often exhibit imbalances in key identity attributes such as gender, age, and ethnicity~\cite{Huber2023BiasAD}. To address this, the FLUXSynID framework allows user-defined attribute classes, each with a set of attributes with associated selection probabilities. Additionally, each class can be assigned an inclusion probability during sample generation. In \cref{fig:document_images_workflow}, the \textit{gender} class is always included (100\%), with the \textit{Female} identity attribute selected 49\% of the time. Other classes, such as \textit{body type}, may be included in only a portion of samples (e.g., 50\%), enabling fine-grained control over attribute distributions.

To maintain semantic consistency in attribute combinations, FLUXSynID implements user-defined attribute clash detection. For example, selecting \textit{Bald} as a hairstyle attribute would automatically exclude hair color attributes, as they are semantically incompatible. This system ensures the generated identity attributes remain logically coherent.

Once attributes are selected, they are stored and passed to a Large Language Model (LLM). We use Qwen2.5~\cite{qwen2.5} to generate natural language prompts that integrate all attributes in a semantically rich format. This approach aligns with FLUX.1 [dev]'s architecture, which leverages a T5~\cite{t5} text encoder to process detailed prompts effectively.

\subsection{Document Image LoRA Adapter} \label{sec:lora}
Low-Rank Adaptation (LoRA)~\cite{lora} enables efficient fine-tuning of large pre-trained models by introducing a small set of trainable parameters. Instead of updating the full model weight matrix $W_0 \in \mathbb{R}^{d \times k}$, the LoRA adapter keeps $W_0$ fixed and learns a low-rank update $\Delta W = BA$, where $A \in \mathbb{R}^{d \times r}$ and $B \in \mathbb{R}^{r \times k}$, with $r \ll \min(d, k)$. The adapted weights become $W = W_0 + \Delta W$. This approach significantly reduces trainable parameters compared to methods such as DreamBooth~\cite{dreambooth}.

\begin{figure}
\centering
\includegraphics[width=\linewidth]{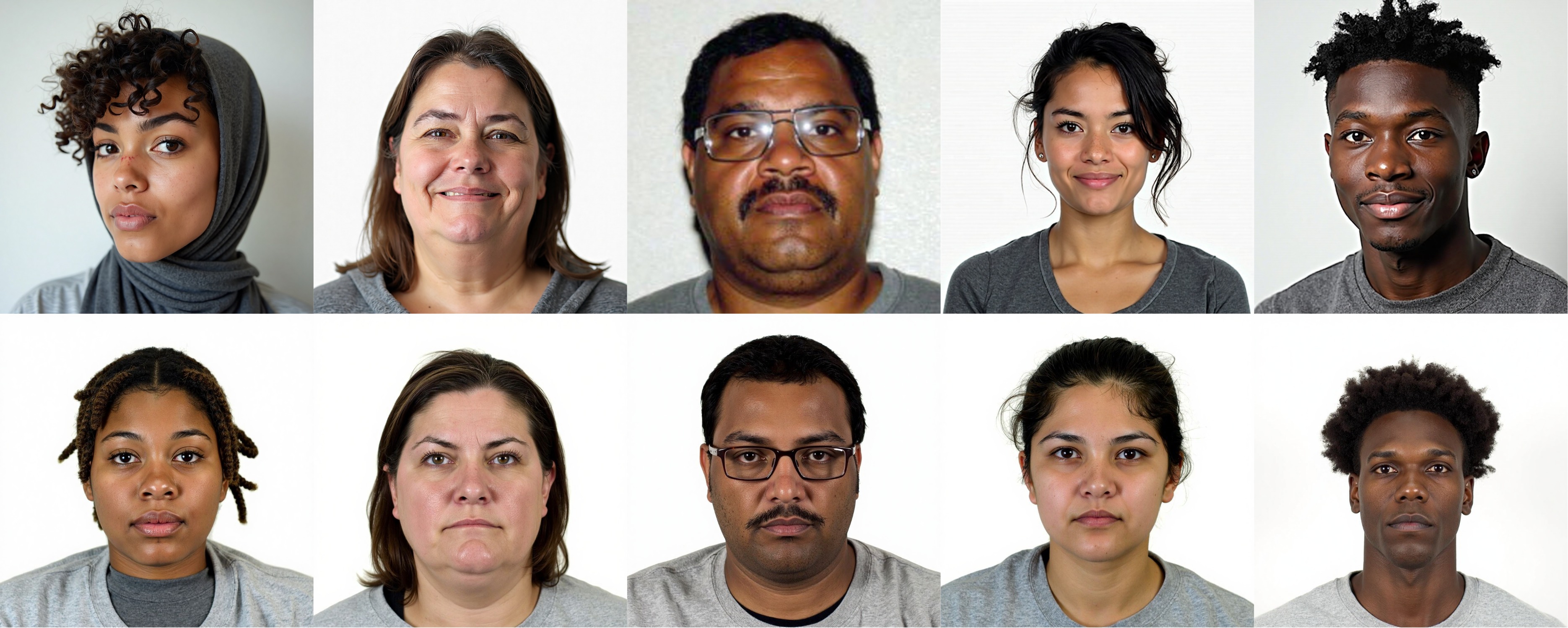}
\caption{(Top row): Synthetic document images generated \textit{without} LoRA~\cite{lora}, exhibiting inconsistent head poses, facial expressions, and lighting. (Bottom row): Images generated \textit{with} LoRA fine-tuning, showing more uniform, document-style outputs with neutral expressions and consistent photographic conditions.}
\label{fig:lora_vs_no_lora}
\end{figure}

We fine-tune the FLUX.1 [dev] diffusion model using a LoRA adapter to ensure stylistic consistency in generating high-resolution, vivid frontal face images with neutral expressions and a uniform white background. The adapter is trained to capture the visual style of images from the Chicago Face Database (CFD)~\cite{chicago_db_1, chicago_db_2, chicago_db_3}. \Cref{fig:lora_vs_no_lora} compares results with and without LoRA, showing improvements in pose, lighting, and expression consistency.

\subsection{Image Generation}
Once trained, the LoRA adapter updates the weights of the pre-trained and frozen FLUX.1 [dev] model, which generates images from prompts produced by Qwen2.5. FLUX.1 [dev] is a latent diffusion model based on a rectified flow transformer~\cite{rectifying_flow_transformer}, designed for text-to-image synthesis. It integrates Classifier-Free Guidance~\cite{cfg} directly into the transformer via a single modulation parameter, the \textit{guidance scale}, eliminating the need for negative prompts and dual-pass inference (i.e., conditional and unconditional passes). This halves inference time and enables direct control over prompt adherence: lower guidance yields more diverse outputs, while higher guidance increases prompt adherence at the cost of variation.

As shown in \cref{fig:document_images_workflow}, the generated prompt $P$ is encoded using the pre-trained and frozen CLIP-L~\cite{clip} and T5-XXL~\cite{t5} models. FLUX.1 [dev] combines these encoders in a complementary way. CLIP's global embedding is fused with the timestep embedding to modulate the transformer layers. Meanwhile, T5-XXL's token embeddings are concatenated with latent image tokens to form a unified input sequence, which is processed via self-attention. This design mimics cross-attention, enabling the text tokens to influence specific regions in the image~\cite{eraseanything}.

After encoding the text prompt $P$ using both encoders ($E_{\text{CLIP-L \& T5-XXL}}$), the resulting textual conditioning $c_P$ is used as conditional input to FLUX.1 [dev] alongside a 16-channel Gaussian noise latent image $z_t$ sampled from a random seed. Through latent diffusion, FLUX.1 [dev] iteratively denoises $z_t$ over $t$ steps to produce the final latent image. The conditional denoising process is defined as:

\begin{equation}
    z_{t-1} = \text{DPM++ 2M}\left (z_t, t, \epsilon_{\theta^{CFD}}\left ( z_t, t, c_P\right )\right),
\end{equation}

\noindent where DPM++ 2M~\cite{dpmpp} is used as the sampling algorithm in diffusion probabilistic modeling, and $\epsilon_{\theta^{CFD}}$ represents the FLUX.1 [dev] model with the LoRA adapter trained on CFD images (see \cref{sec:lora}).

The final denoised latent image, $z_0$, is then passed to the pre-trained FLUX.1 [dev] Variational Autoencoder (VAE) decoder network~\cite{flux}, $D_{\text{VAE}}$, which generates the synthetic document image, $I_D$, based on the textual prompt.

\begin{figure*}
\centering
\includegraphics[width=0.97\linewidth]{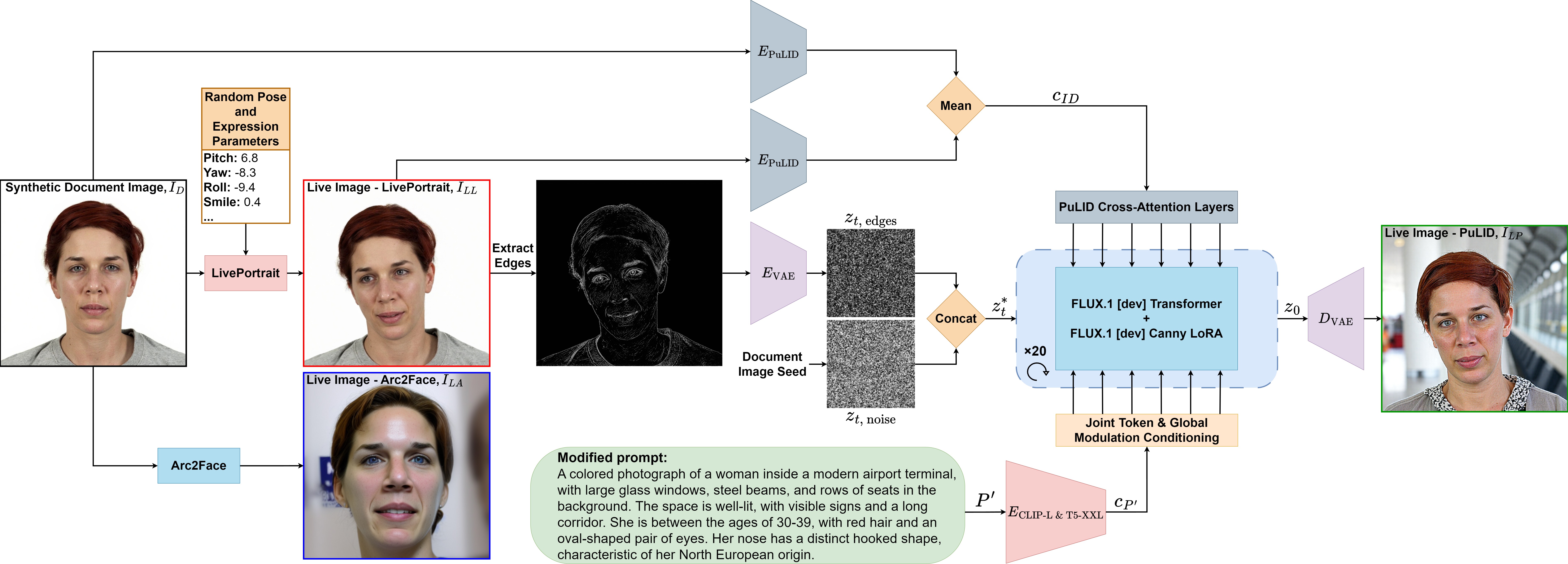}
\caption{Overview of the FLUXSynID live capture image generation pipeline. Starting from a synthetic document image, three methods are used to produce identity-consistent live images: LivePortrait~\cite{live_portrait} enables controlled variations in pose and expression; Arc2Face~\cite{arc2face} introduces natural diversity via identity embeddings; and the PuLID-based approach~\cite{pulid, pulid_github} is integrated with FLUX.1 [dev]~\cite{flux} via identity and edge-based conditioning, guided by modified image prompts. Please refer to \cref{sec:live_imgs} for details.}
\label{fig:live_images_workflow}
\end{figure*}

\section{FLUXSynID: Trusted Live Capture Images} \label{sec:live_imgs}

To reflect real-world identity verification scenarios (e.g., airport gates) in the FLUXSynID framework, we incorporate three complementary generative approaches: LivePortrait~\cite{live_portrait}, Arc2Face~\cite{arc2face}, and PuLID~\cite{pulid}. As illustrated in \cref{fig:live_images_workflow}, LivePortrait and Arc2Face are applied in their original forms, providing precise control and natural variability, respectively. In contrast, our primary contribution lies in the extensive adaptation of PuLID through novel conditioning strategies that enhance pose and expression diversity while preserving identity. Together, these methods generate live-style variants $I_{LL}$, $I_{LA}$, and $I_{LP}$, respectively, from the original synthetic document images $I_D$.

\subsection{LivePortrait-based Image Generation}
LivePortrait~\cite{live_portrait} is a facial animation method that leverages implicit 3D keypoints to control expression and head pose from a single image. In FLUXSynID, it is applied to each synthetic document image $I_D$ to generate identity-preserving live capture variants $I_{LL}$ by applying subtle, randomly sampled changes in pose and expression.

\subsection{Arc2Face-based Image Generation}
Arc2Face~\cite{arc2face} generates identity-preserving facial images via ArcFace~\cite{arcface} identity embeddings, which condition the Stable Diffusion~\cite{latent_diffusion} model. The authors report superior identity similarity compared to other generative models, underscoring its effectiveness. Notably, Arc2Face requires no textual prompts, relying entirely on image-based identity embeddings. In FLUXSynID, Arc2Face is applied to the synthetic document images $I_D$ to produce live-style variants $I_{LA}$, complementing LivePortrait by contributing unstructured variability. This variability is driven solely by identity conditioning and helps balance LivePortrait's more controlled outputs.

\subsection{PuLID-based Image Generation}
PuLID (Pure and Lightning ID Customization)~\cite{pulid} personalizes text-to-image diffusion models using identity-specific embeddings derived from reference images. Similar to Arc2Face, PuLID enables rapid identity injection without requiring network retraining for each subject.

In FLUXSynID, we employ PuLID-FLUX-v0.9.1~\cite{pulid_github}, an adaptation of PuLID integrated with the pre-trained and frozen FLUX.1 [dev]~\cite{flux} diffusion model. As shown in \cref{fig:live_images_workflow}, synthetic document images ($I_D$) and LivePortrait-modified images ($I_{LL}$) are passed through PuLID's identity encoder to extract static and dynamic identity features, which are averaged into a conditioning signal $c_{ID}$.

PuLID introduces dedicated cross-attention layers into FLUX.1 [dev]'s transformer blocks to integrate $c_{ID}$. These cross-attention layers inject identity information directly into the model by enabling focused interactions between the identity representation and the image tokens. This contrasts with the integration of text embeddings, which are incorporated via global modulation and token concatenation.

Simultaneously, the original identity prompt $P$ is transformed into a contextual prompt $P'$, describing the same individual in an airport setting, and encoded via CLIP-L~\cite{clip} and T5-XXL~\cite{t5} to produce the textual condition $c_{P'}$. To enhance structural fidelity, edges are extracted from $I_{LL}$ to provide spatial guidance. These edges are encoded into a latent image $z_{t, \text{edges}}$ using FLUX.1 [dev] VAE encoder~\cite{flux} and are then concatenated along the channel dimension with a Gaussian noise latent image $z_{t, \text{noise}}$ to form $z_t^*$.


The concatenated latent image $z_t^*$ serves as the input for the FLUX.1 [dev] diffusion model equipped with a specialized FLUX.1 [dev] Canny LoRA adapter~\cite{flux}. This LoRA adapter modifies the input layer of FLUX.1 [dev] and is trained for edge-based image-to-image conditioning.

The final generation is conditioned on three signals: identity embedding ($c_{ID}$), modified text embedding ($c_{P'}$), and latent edge representation ($z_{t, \text{edges}}$). This ensures identity consistency while introducing natural pose and expression variations. The denoising process is defined as:

\begin{equation}
z_{t-1}^* = \text{DPM++ 2M}\left ( z_t^*, t, \epsilon_{\theta^{\text{Canny}}}\left ( z_t^*, t, c_{P'}, c_{\text{ID}}\right )\right ),
\end{equation}

\noindent where $\epsilon_{\theta^{Canny}}$ represents FLUX.1 [dev] model combined with the FLUX.1 [dev] Canny LoRA adapter.

\section{Experiments}
\subsection{Implementation Details}
\noindent\textbf{Prompt Generation.} Identity prompts were generated using the Qwen2.5~\cite{qwen2.5} LLM, based on sampled identity attributes. Attribute inclusion probabilities and conflict rules were applied to generate semantically rich prompts optimized for conditioning FLUX.1 [dev]~\cite{flux}.

\noindent\textbf{LoRA Training for Document Image Generation.} A LoRA adapter was trained using FluxGym~\cite{fluxgym} on 830 images from the Chicago Face Database (CFD)~\cite{chicago_db_1, chicago_db_2, chicago_db_3}, which provides high-resolution, front-facing portraits with neutral expressions and consistent lighting. The adapter was trained with rank $r=16$, learning rate 0.0002, and 18 training epochs. Image captions were generated using the InternVL2.5~\cite{InternVL2_5} vision-language model. The LoRA module was activated during inference via a unique token (``s7ll2f4h"), and additional parameters in the CLIP-L~\cite{clip} encoder were also fine-tuned.

\noindent\textbf{Document Image Generation.} Document images were synthesized through the ComfyUI-based~\cite{comfyui} implementation of FLUX.1 [dev]~\cite{flux}, conditioned on Qwen2.5 prompts. Guidance scales were sampled randomly between 1.7 and 2.5 to balance fidelity with identity diversity. Each image was generated using $t = 20$ diffusion steps.

\noindent\textbf{Live Image Generation.} LivePortrait~\cite{live_portrait} and PuLID-FLUX-v0.9.1~\cite{pulid_github} were used via modified ComfyUI-based~\cite{comfyui} implementations. A fixed guidance scale of 4 and $t = 20$ steps were used with PuLID, as recommended by the authors. Arc2Face~\cite{arc2face} was run via the official implementation with default hyperparameters.

\noindent\textbf{Dataset Composition.} A total of 15,000 synthetic identities were generated across 14 attribute classes: \textit{gender}, \textit{age}, \textit{region of origin}, \textit{body type}, \textit{eye shape}, \textit{lip shape}, \textit{nose shape}, \textit{face shape}, \textit{hairstyle}, \textit{hair color}, \textit{eyewear}, \textit{facial hair}, \textit{skin type}, and \textit{ICAO-compliant headwear}~\cite{icao}. The \textit{gender}, \textit{age}, and \textit{region of origin} attributes were included in 100\% of prompts to ensure consistent representation. The distributions for \textit{age}, \textit{region of origin}, and \textit{body type} were uniform. The remaining attributes were given heuristic probabilities based on rough estimates of broad, plausible trends (e.g., certain facial hair styles are naturally less common than a clean-shaven look). In the absence of comprehensive demographic data, these estimates were used to illustrate the dataset generation process. Researchers are encouraged to define custom distributions for their specific needs.

Of the 15,000 identities, 111 were excluded due to Arc2Face~\cite{arc2face} failing to extract identity embeddings. All images were generated at a resolution of $1024 \times 1024$, except for Arc2Face-based live images, which were constrained to $512 \times 512$ due to implementation limitations.

\subsection{Similarity-based Identity Filtering} \label{sec:similarity_filtering}
Due to prompt similarity and inherent biases in the FLUX.1 [dev]~\cite{flux} model, some synthetic identities may appear overly similar. Unlike ONOT~\cite{onot}, which enforces strict identity separation by removing all similar identities detected by a Face Recognition System (FRS), FLUXSynID allows a small degree of similarity, reflecting real-world datasets where a limited False Match Rate (FMR) is acceptable. For instance, FRONTEX guidelines~\cite{frontex} permit an FMR of 0.1\%. As the dataset size grows, the probability that a randomly selected identity $i$ falsely matches at least one other identity in the dataset increases, and is defined as:

\begin{equation}
P_{\text{False Match}}(i) = 1 - (1 - \text{FMR})^{N-1},
\label{eq:p_false_match}
\end{equation}

\noindent where $N$ represents the total number of identities.

To manage this behavior, FLUXSynID employs iterative filtering based on FRS similarity scores. Starting from an initial set of document images $D_o$, we identify inter-identity pairs that are falsely matched. The filtering process iteratively removes identities involved in the highest number of such matches. This continues until the dataset-wide FMR reaches or falls below a predefined static FMR target:

\begin{equation}
\text{FMR}_{D_f} \leq \text{FMR}_{\text{target}}.
\label{eq:fmr_d_f}
\end{equation}

Here, dataset-wide FMR is calculated as the ratio of the number of pairs exceeding the threshold to the total number of possible comparisons. This method ensures the final subset $D_f \subseteq D_o$ maintains a practical and acceptable FMR.

Experiments in this study leverage two open-source face recognition models: ArcFace~\cite{arcface} and AdaFace~\cite{adaface}. For each, two decision thresholds are selected, corresponding to FMR targets of 0.1\% and 0.01\%. These thresholds are derived from 340k impostor verification attempts on identities from the CFD dataset, yielding values of 0.423 and 0.497 for ArcFace, and 0.253 and 0.334 for AdaFace, respectively. Accordingly, the $\text{FMR}_{\text{target}}$ in Equation~\eqref{eq:fmr_d_f} is set to 0.1\% or 0.01\%, depending on the chosen threshold.

The number of identities remaining after filtering varies significantly depending on the FRS used and the FMR target. \Cref{tab:fmr_filtering_results} summarizes the results across both ArcFace and AdaFace for target FMRs of 0.1\% and 0.01\%.

\begin{table}[htbp]
\centering
\begin{tabular}{lcc}
\hline
\textbf{FRS} & \textbf{FMR Target} & \textbf{Remaining Identities (\%)} \\
\hline
ArcFace & 0.01\% & 9{,}358 (62.9\%) \\
ArcFace & 0.1\%  & 6{,}641 (44.6\%) \\
\hline
AdaFace & 0.01\% & 6{,}074 (40.8\%) \\
AdaFace & 0.1\%  & 2{,}389 (16.0\%) \\
\hline
\end{tabular}
\caption{Remaining identities after similarity-based filtering at different FMR targets using ArcFace~\cite{arcface} and AdaFace~\cite{adaface}. Percentages are relative to the initial 14{,}889 identities. Note: lower FMR targets require higher FRS decision thresholds, reducing the number of identity pairs flagged as false matches. Thus, fewer identities are removed and a larger portion of the dataset is retained.}
\label{tab:fmr_filtering_results}
\end{table}

For comparison, we also applied ONOT's~\cite{onot} stricter identity filtering, which removes all identity pairs exceeding the threshold, effectively enforcing a dataset-wide FMR of zero. Using ArcFace with ONOT's specified thresholds, FLUXSynID retained 51.0\% of identities at FMR 0.1\% and 87.1\% at FMR 0.01\%. In contrast, ONOT reports retaining only 3.1\% and 6.3\%, respectively. Although the comparison is not exact due to ONOT's use of additional intra-class filtering, these results suggest that FLUXSynID identities exhibit substantially greater feature-level diversity.

Finally, to ensure there is no identity leakage from LoRA fine-tuning on CFD images, we compared all 14,889 FLUXSynID identities against the 830 CFD subjects used for training. Using AdaFace and a threshold of 0.539, corresponding to the minimum similarity score among mated pairs in the FRGC~\cite{frgc} dataset, no matches were found. This confirms that no identity leakage occurred.

\subsection{Evaluation}
\begin{figure*}
\centering
\includegraphics[width=0.87\linewidth]{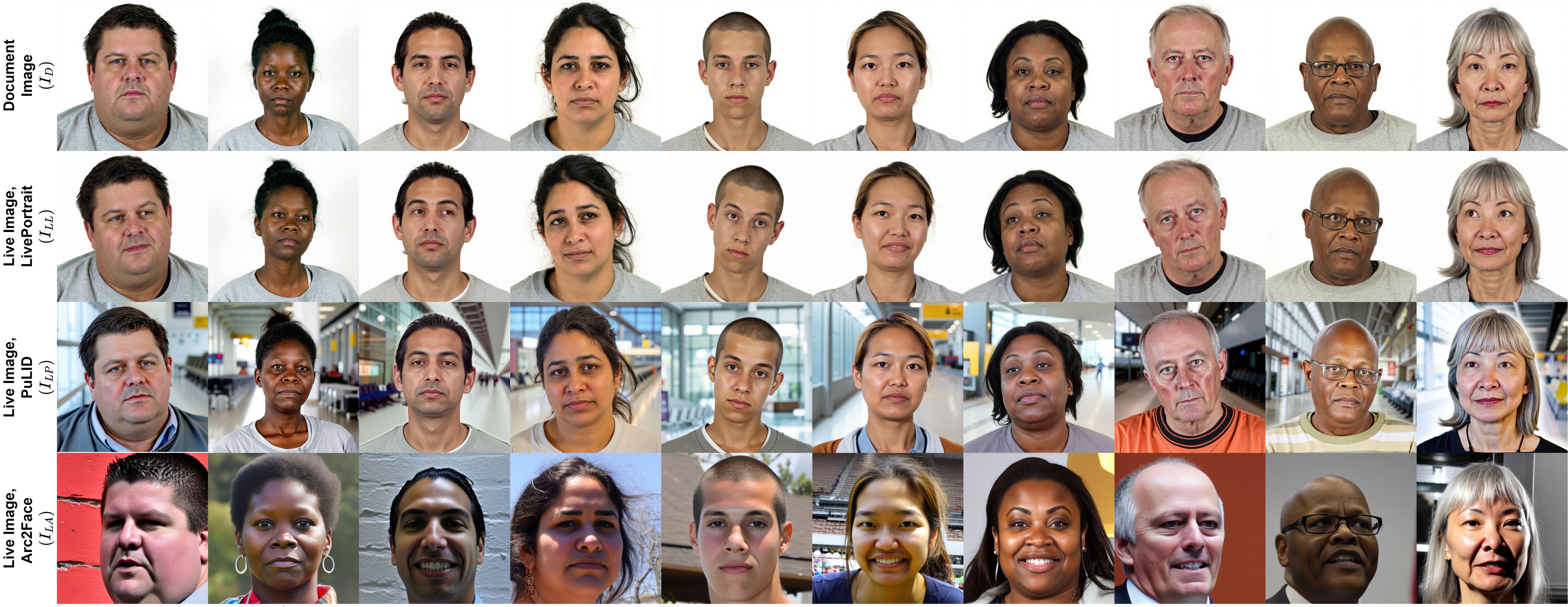}
\caption{Examples of synthetic identities from FLUXSynID, each shown with a document image ($I_D$) and corresponding live capture images generated using LivePortrait ($I_{LL}$)~\cite{live_portrait}, PuLID ($I_{LP}$)~\cite{pulid}, and Arc2Face ($I_{LA}$)~\cite{arc2face}.}
\label{fig:visual_examples}
\end{figure*}

\noindent\textbf{Visual Results.} \Cref{fig:visual_examples} shows sample identities from FLUXSynID, each consisting of a document image ($I_D$) and live capture images generated by LivePortrait ($I_{LL}$), PuLID ($I_{LP}$), and Arc2Face ($I_{LA}$). Document images are frontal, display neutral expressions, and have uniform white backgrounds, reflecting stylistic control via CFD-based LoRA fine-tuning.

Each live image generation method introduces different levels of variation. LivePortrait introduces subtle, controlled changes in pose and expression while maintaining consistent lighting and background, although minor blurring is often evident. PuLID leverages text, edge maps, and identity embeddings to alter both background and pose, although it may occasionally cause identity drift. Arc2Face generates the most diverse outputs, introducing substantial changes in setting, lighting, and expression that closely mimic in-the-wild photographs.

\begin{figure}
\centering
\includegraphics[width=0.9\linewidth]{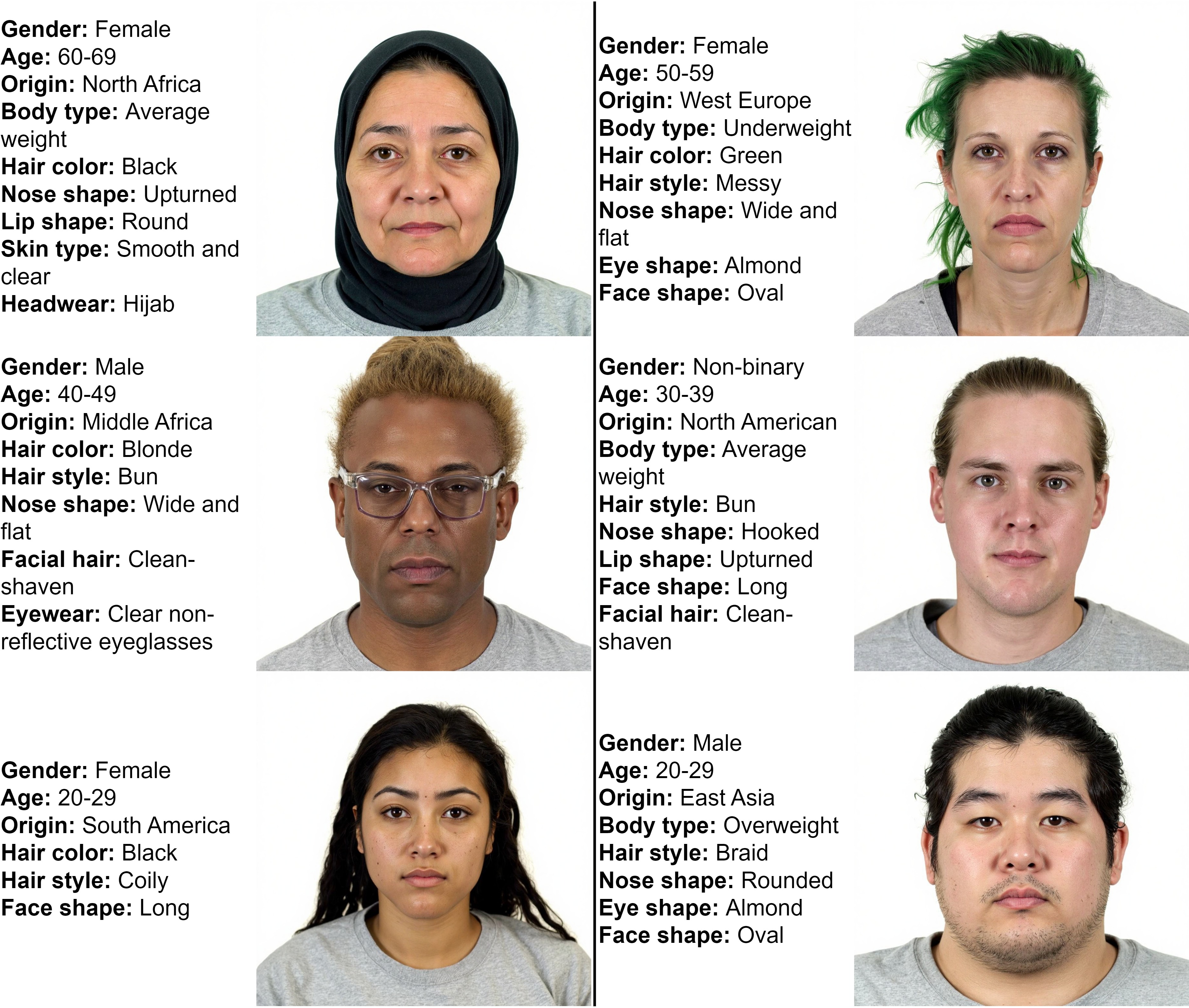}
\caption{Examples of randomly sampled identity attributes and their corresponding synthetic document images ($I_D$) generated using the FLUXSynID framework.}
\label{fig:prompt-to-image}
\end{figure}

\noindent\textbf{Prompt-Image Consistency.} \Cref{fig:prompt-to-image} shows synthetic document images generated from randomly sampled identity attributes. Generally, the images align well with the specified attributes, although some limitations are evident. Age estimation occasionally deviates, producing individuals who appear significantly younger or older than intended, consistent with findings in~\cite{flux_age_estimation}, which highlight FLUX.1 [dev]'s difficulty in accurately representing age.

Subtle facial features, such as nose shape, are also inconsistently realized. This may stem from limitations in automated image captioning, which often omits fine-grained details when annotating datasets for training text-conditioned diffusion models~\cite{Kreiss2021ConcadiaTI, Dess2023CrossDomainIC, Fisch2020CapWAPIC}.

Empirically, guidance scales between 1.7 and 2.5 yielded the best trade-off between fidelity and prompt consistency. Below 1.7, images often exhibited artifacts and poor prompt adherence, while scales above 2.5 produced overly saturated and visually similar faces, increasing removals during similarity-based filtering.

\begin{figure*}
\centering
\includegraphics[width=0.99\linewidth]{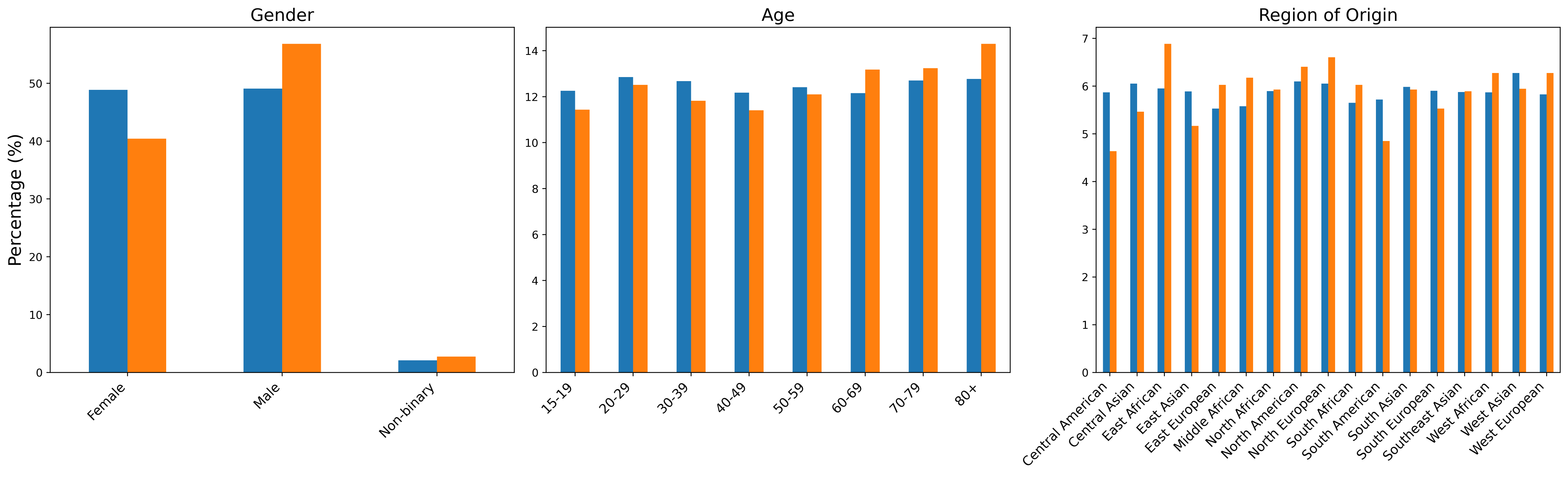}
\caption{Distribution of identity attributes in the dataset before and after filtering based on identity similarity. Each bar plot shows the percentage of images with a given attribute. Blue bars represent the original dataset, and orange bars represent the dataset after filtering.}
\label{fig:bar_plots}
\end{figure*}

\noindent\textbf{Impact of Similarity-based Filtering on Attribute Distributions.} To evaluate the potential bias introduced by similarity-based filtering (\cref{sec:similarity_filtering}), \cref{fig:bar_plots} compares the distributions of \textit{gender}, \textit{age}, and \textit{region of origin} attributes before and after filtering.

Female identities were observed to be removed more frequently, reducing their overall representation. This indicates that FLUX.1 [dev] generates less diverse female faces, leading to higher similarity scores and more frequent exclusion. Conversely, male and non-binary identities increased proportionally, indicating greater variability.

In the \textit{age} category, younger individuals were more frequently filtered out, reflecting reduced variation among these groups. Attribute distributions for \textit{region of origin} remained largely stable, although some shifts were observed. For example, there was a decline in Central American identities and an increase in East African ones, highlighting uneven feature diversity across regions that may stem from biases in the generative model.

\begin{figure}[htbp]
\centering
\begin{subfigure}{0.49\linewidth}
    \includegraphics[width=\linewidth]{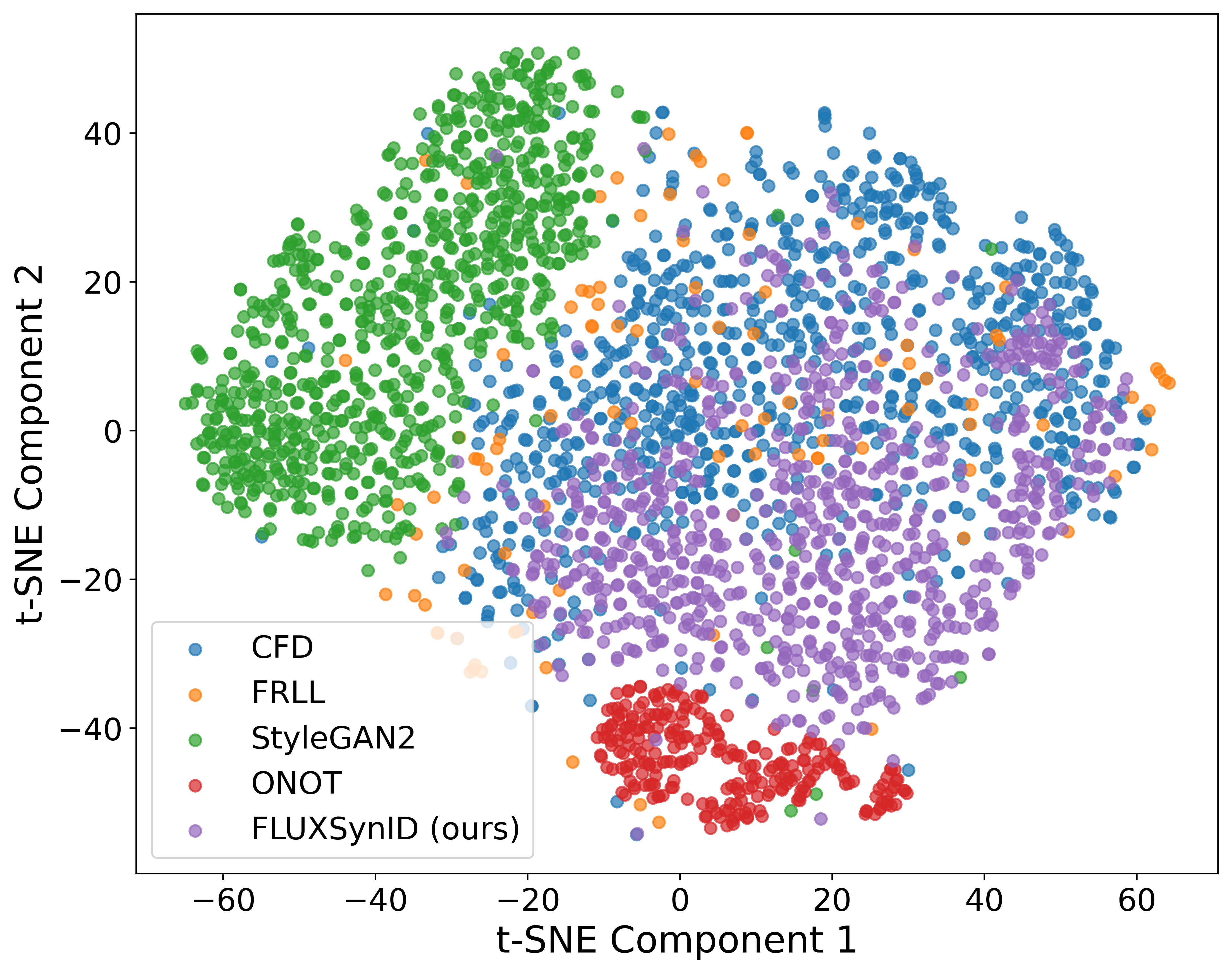}
    \caption{ArcFace~\cite{arcface}}
    \label{fig:tsne_arcface}
\end{subfigure}
\hfill
\begin{subfigure}{0.49\linewidth}
    \includegraphics[width=\linewidth]{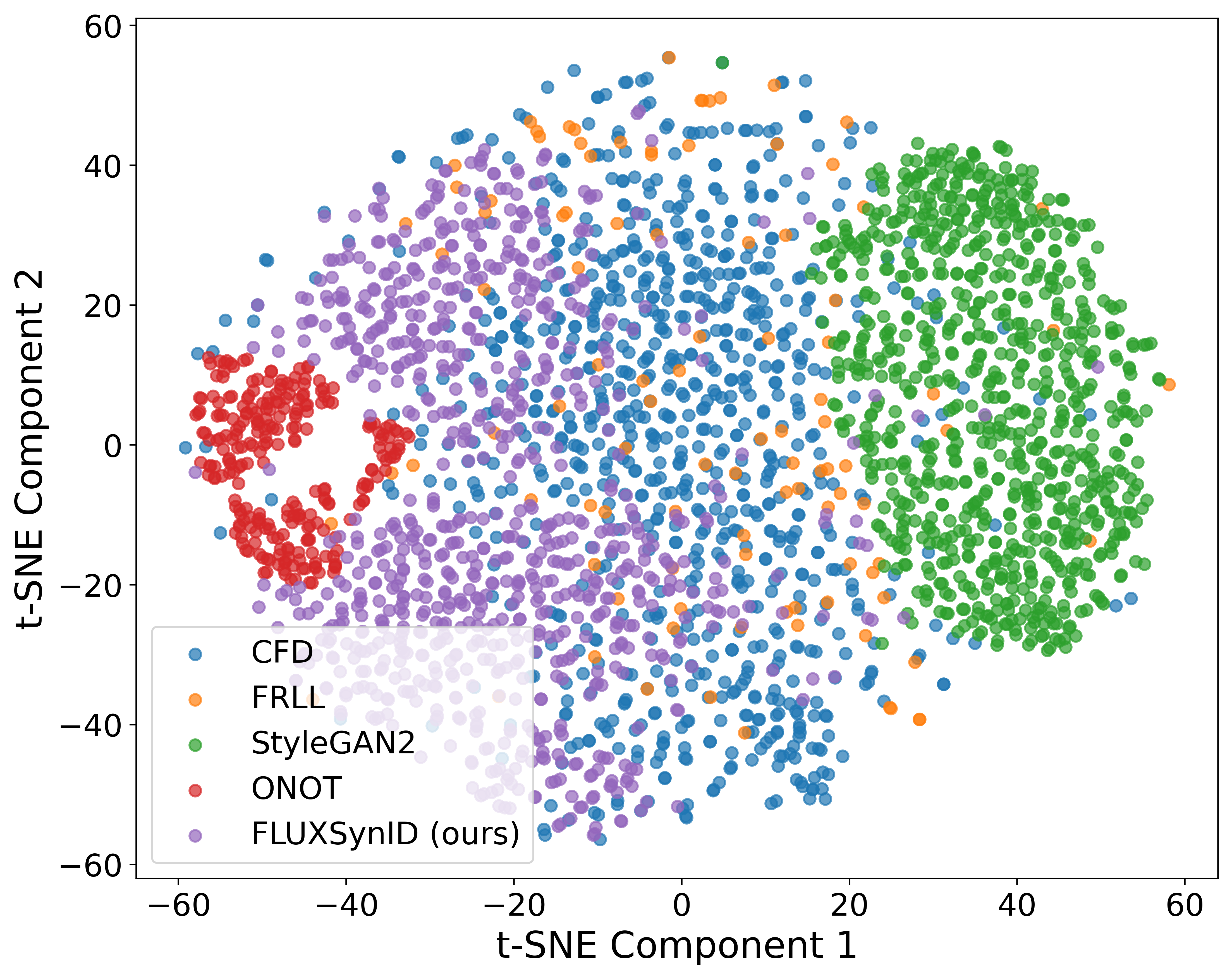}
    \caption{AdaFace~\cite{adaface}}
    \label{fig:tsne_adaface}
\end{subfigure}
\caption{t-SNE~\cite{tsne} visualization of FRS embeddings derived from real and synthetic identity images.}
\label{fig:tsne}
\end{figure}

\noindent\textbf{Distribution of Identity Embeddings.} \Cref{fig:tsne} visualizes identity embeddings extracted by ArcFace~\cite{arcface} and AdaFace~\cite{adaface} across real and synthetic datasets. Synthetic StyleGAN2~\cite{stylegan2} images were generated using the pSp~\cite{psp} encoder, producing random identities in frontal poses. For ONOT~\cite{onot}, ICAO-compliant images from Subset 1 (the largest identity set) were used. FLUXSynID identities were filtered using FMR of 0.01\% thresholds (\cref{sec:similarity_filtering}) to remove highly similar identities. Only document images were considered to maintain consistency across datasets, all of which contain document-like image types.

The results show that StyleGAN2 and ONOT identities form distinct clusters that do not overlap well with real-world identity distributions from CFD~\cite{chicago_db_1, chicago_db_2, chicago_db_3} and FRLL~\cite{frll}. In contrast, FLUXSynID embeddings demonstrate greater overlap with real data, suggesting closer alignment with real-world identity variability. Imperfect overlap may reflect FLUXSynID's broader demographic range, particularly in age and region of origin, compared to the more limited scope of CFD and FRLL.

Moreover, ONOT identities form tight clusters, indicating low inter-class diversity. In contrast, FLUXSynID embeddings are more evenly distributed, capturing a broader and more realistic spectrum of identity features.

\begin{figure}[htbp]
\centering

\begin{subfigure}{0.32\linewidth}
    \includegraphics[width=\linewidth]{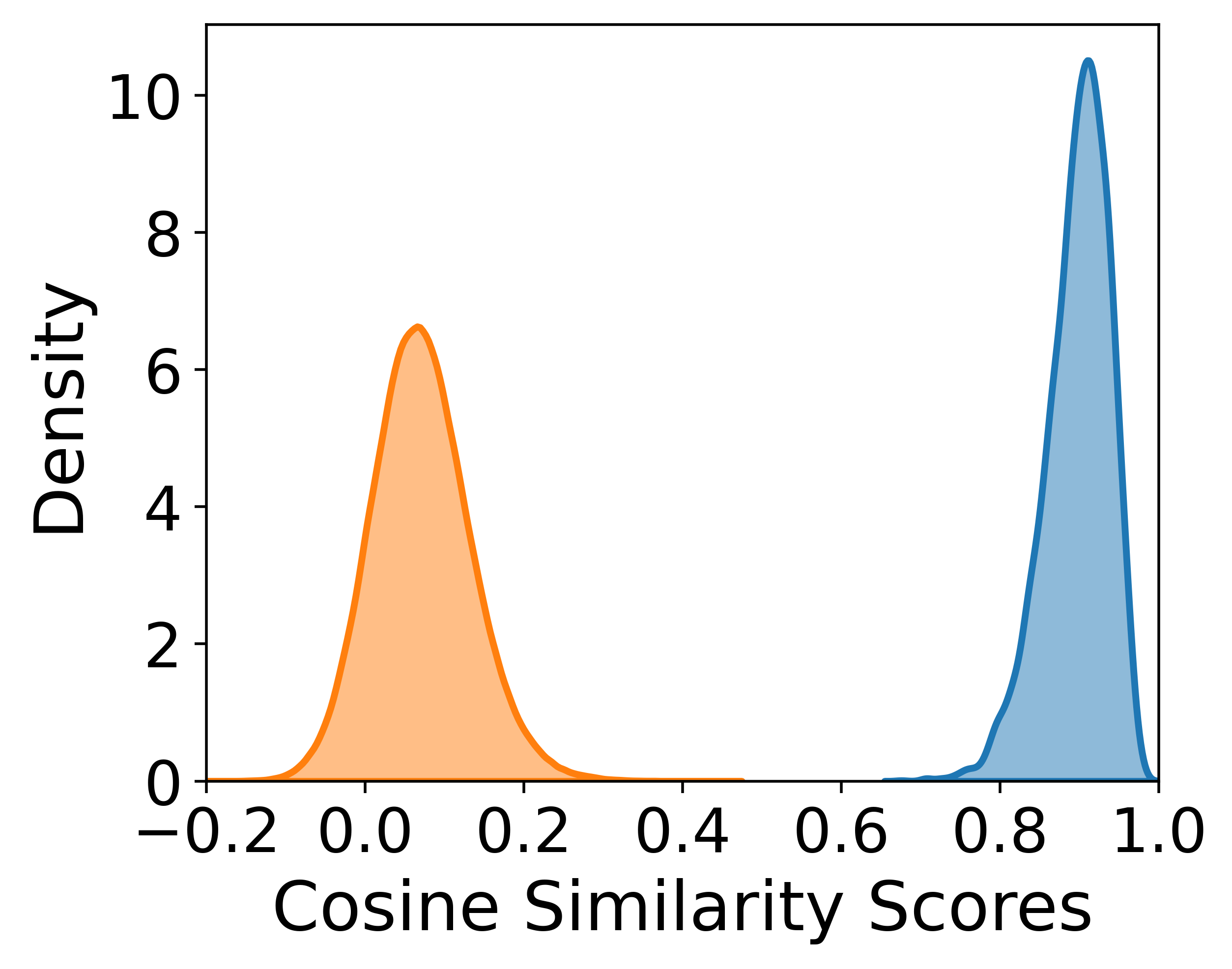}
    \caption{FLUXSynID-LL}
    \label{fig:hist_FLL}
\end{subfigure}
\hfill
\begin{subfigure}{0.32\linewidth}
    \includegraphics[width=\linewidth]{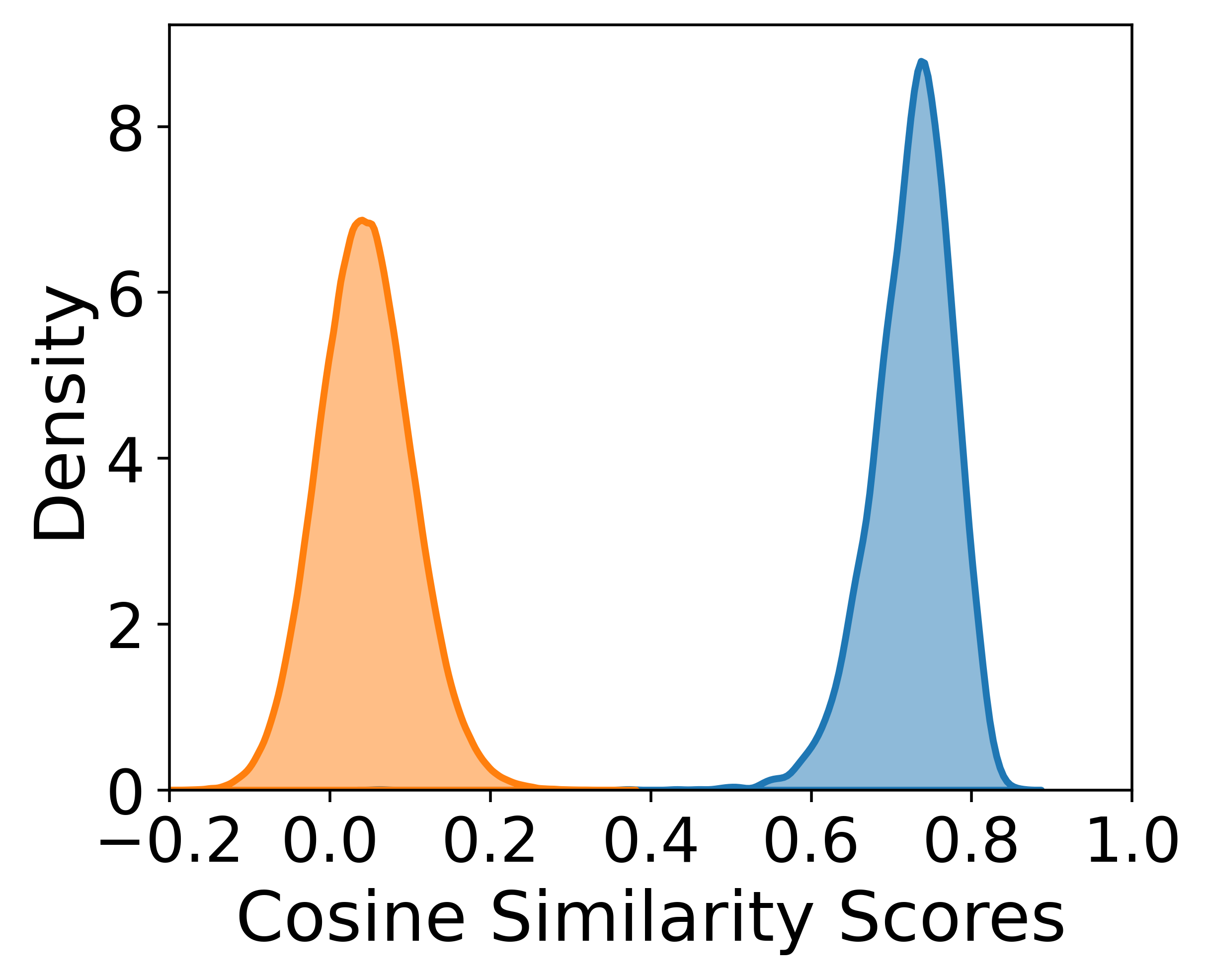}
    \caption{FLUXSynID-LA}
    \label{fig:hist_FLA}
\end{subfigure}
\hfill
\begin{subfigure}{0.32\linewidth}
    \includegraphics[width=\linewidth]{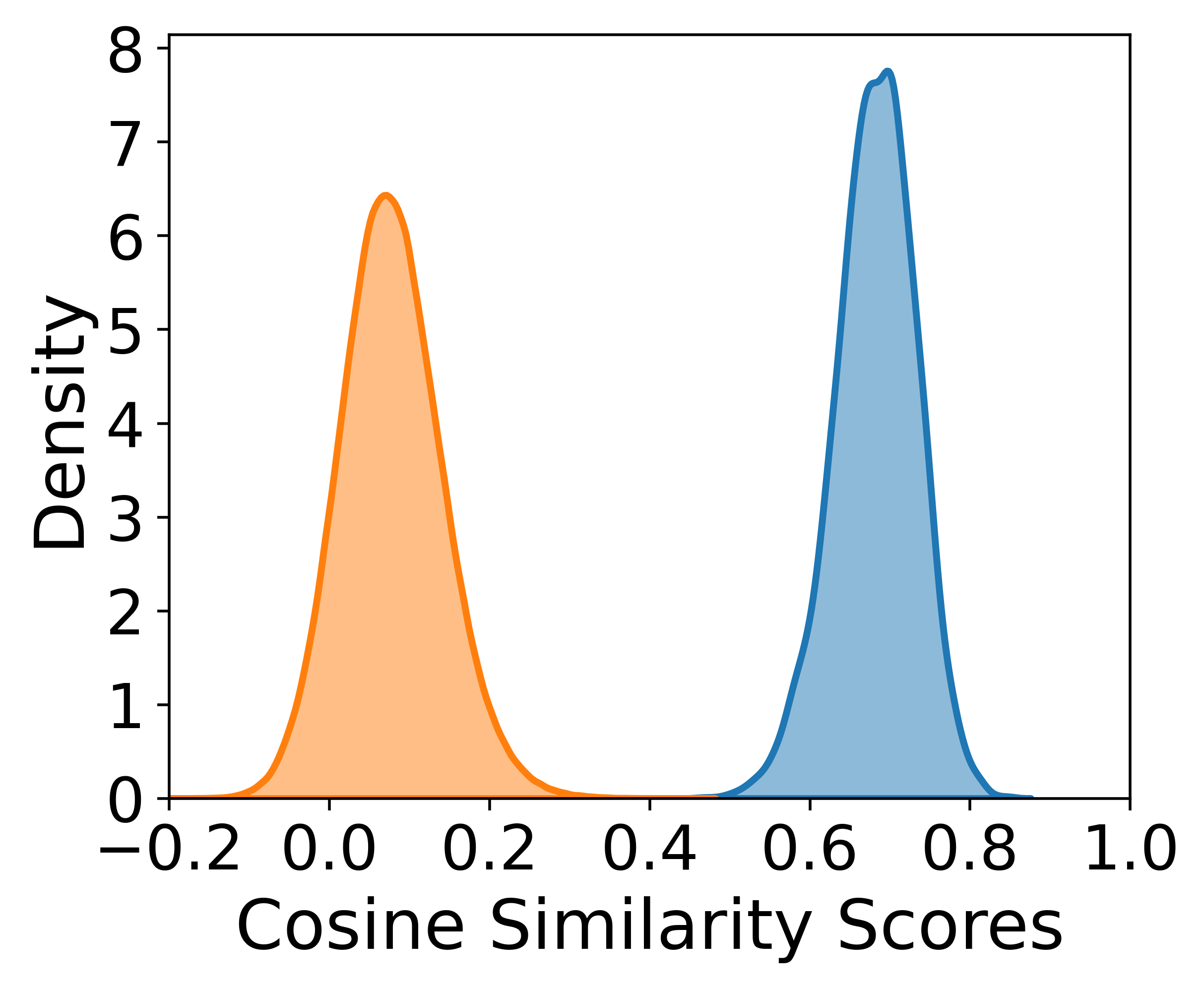}
    \caption{FLUXSynID-LP}
    \label{fig:hist_FLP}
\end{subfigure}

\vspace{0.5em} 

\begin{subfigure}{0.32\linewidth}
    \includegraphics[width=\linewidth]{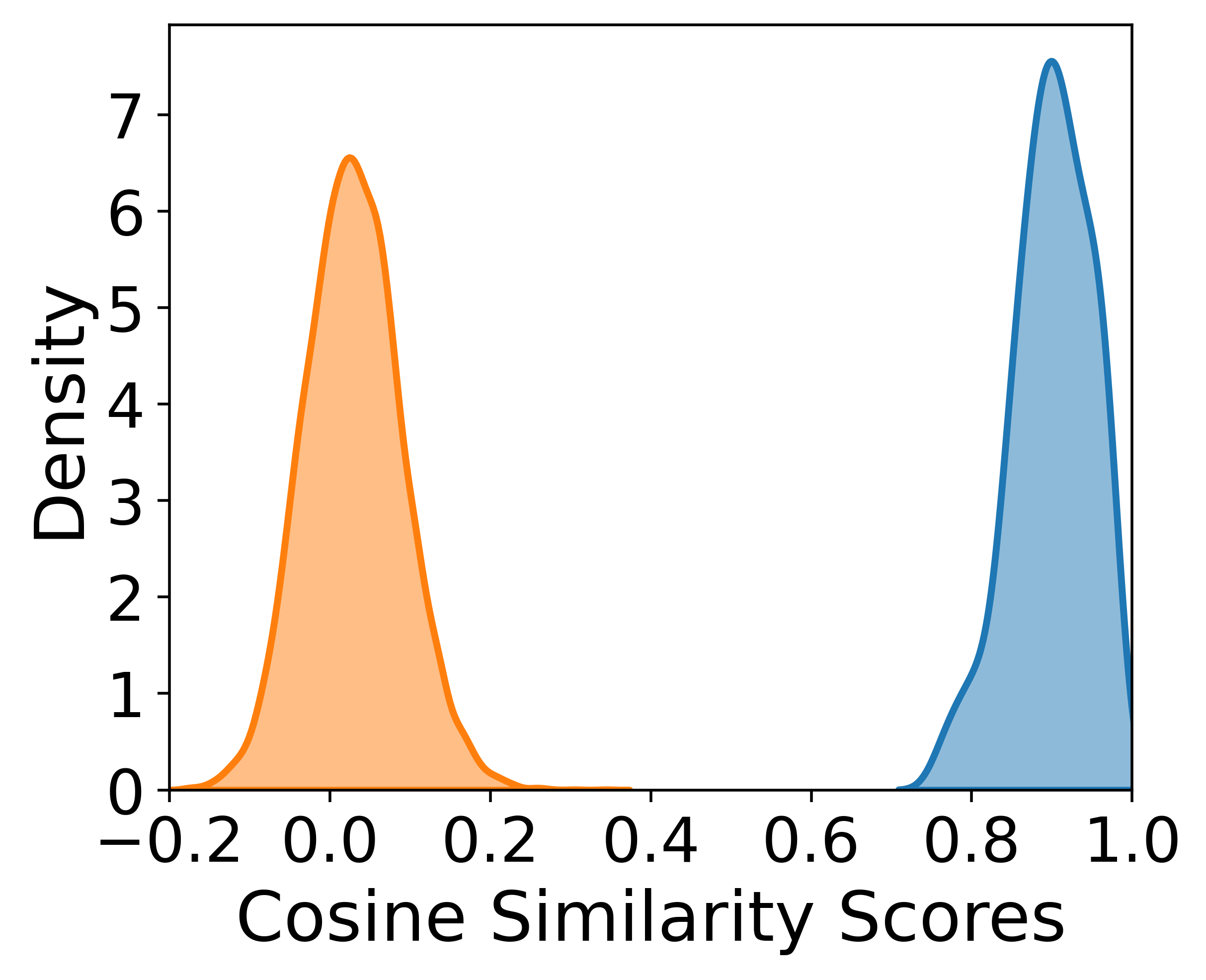}
    \caption{FRLL~\cite{frll}}
    \label{fig:hist_FRLL}
\end{subfigure}
\hfill
\begin{subfigure}{0.32\linewidth}
    \includegraphics[width=\linewidth]{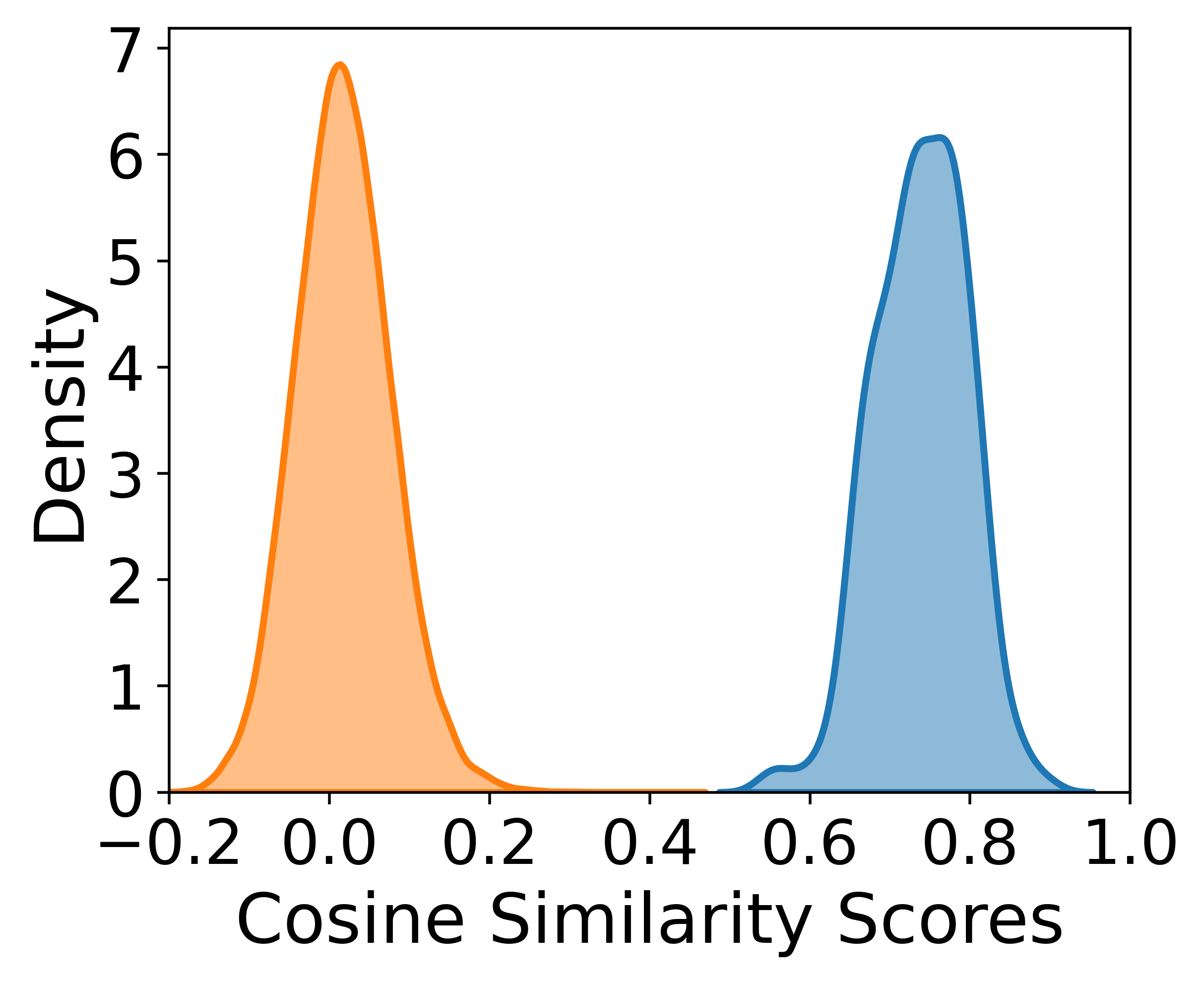}
    \caption{FRGC~\cite{frgc}}
    \label{fig:hist_FRGC}
\end{subfigure}
\hfill
\begin{subfigure}{0.32\linewidth}
    \includegraphics[width=\linewidth]{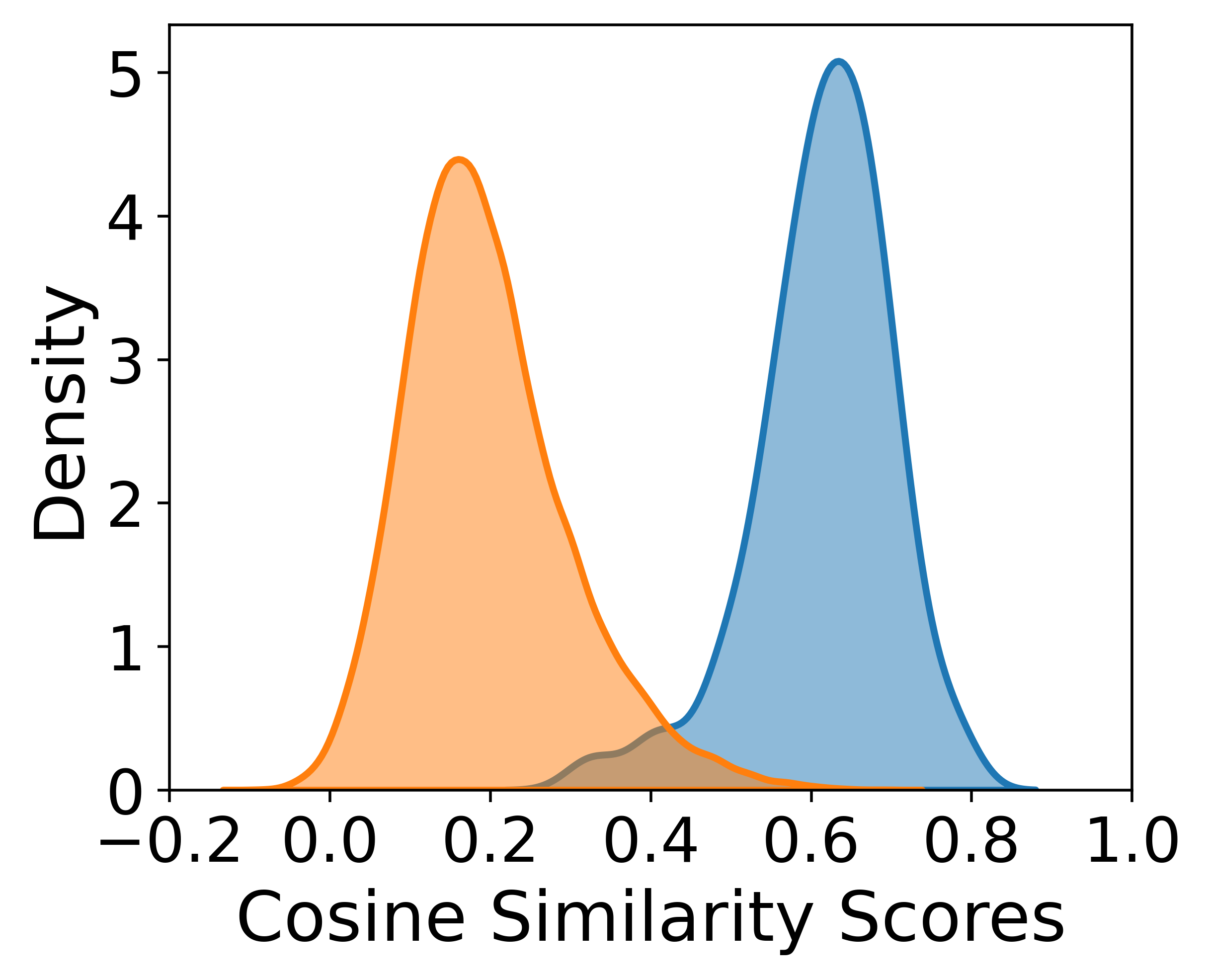}
    \caption{ONOT~\cite{onot}}
    \label{fig:hist_ONOT}
\end{subfigure}
\caption{Distribution of cosine similarity scores for mated (blue) and non-mated (orange) identity pairs across real and synthetic datasets, using the AdaFace~\cite{adaface} FRS with an FMR of 0.01\% threshold. FLUXSynID-LL, -LP, and -LA refer to live images generated using LivePortrait, PuLID, and Arc2Face, respectively.}
\label{fig:hists}
\end{figure}

\begin{table}[htbp]
\centering
\begin{tabular}{lcccc}
\hline
\textbf{Dataset} & \multicolumn{2}{c}{\textbf{Mated}} & \multicolumn{2}{c}{\textbf{Non-Mated}} \\
\cline{2-5}
& FRLL & FRGC & FRLL & FRGC \\
\hline
ONOT & 23.793 & 6.770 & 5.202 & 4.715 \\
FLUXSynID-LL & \textbf{1.047} & 15.512 & 0.248 & 0.375 \\
FLUXSynID-LP & 23.992 & 1.655 & 0.332 & 0.477 \\
FLUXSynID-LA & 19.225 & \textbf{0.145} & \textbf{0.038} & \textbf{0.101} \\
\hline
\end{tabular}
\caption{KL divergence between similarity score distributions of synthetic and real datasets for mated and non-mated identity pairs. Lower values indicate greater distributional similarity. FLUXSynID-LL, -LP, and -LA refer to live images generated using LivePortrait, PuLID, and Arc2Face, respectively.}
\label{tab:kl_divergence}
\end{table}

\noindent\textbf{Similarity Score Distributions.} To further evaluate FLUXSynID, cosine similarity scores for mated (document and corresponding live images) and non-mated pairs (document images compared to live images from 100 other identities) are visualized as histograms in \cref{fig:hists}, with comparisons to real-world FRLL~\cite{frll} and FRGC~\cite{frgc} datasets.

Results indicate that FLUXSynID-LL (LivePortrait) closely matches FRLL, reflecting high identity consistency with minimal appearance variation. FLUXSynID-LP (PuLID) and FLUXSynID-LA (Arc2Face) better align with FRGC, reflecting broader visual diversity. ONOT exhibits overlapping mated and non-mated distributions that are skewed relative to real data, indicating limited identity separation and low inter-class variability.

These observations are quantified in \cref{tab:kl_divergence}, which reports KL divergence between synthetic and real datasets. FLUXSynID-LL yields the lowest mated-pair divergence from FRLL, supporting its close visual match and controlled variation. For non-mated pairs, FLUXSynID-LA achieves the lowest divergence, though its mated-pair scores are higher. Overall, FLUXSynID-LL offers the most balanced similarity to FRLL across both pair types.

For FRGC, FLUXSynID-LA records the lowest mated-pair KL divergence, although FLUXSynID-LP visually resembles FRGC images more closely. While its KL values are slightly higher than Arc2Face's, they still indicate strong distributional similarity.

In contrast, ONOT exhibits high KL divergence across all comparisons, indicating that its similarity score distributions differ substantially from real datasets. This aligns with \cref{fig:hists}, where ONOT's mated and non-mated distributions are clearly shifted relative to real data.

\section{Conclusion}
We present FLUXSynID, a synthetic face dataset generation framework capable of producing high-resolution, identity-consistent document and live capture images with control over identity attributes. Combining prompt-driven diffusion synthesis with LoRA fine-tuning~\cite{lora} and three complementary live image generation methods, FLUXSynID facilitates both controlled and natural variations across identity instances. Experimental results show that FLUXSynID aligns more closely with real-world data distributions than existing synthetic alternatives, while its controlled diversity supports realistic biometric evaluation. These results indicate that FLUXSynID can support biometric tasks requiring structured, identity-consistent image pairs, such as D-MAD, face recognition, and emotion recognition.

Despite its strengths, FLUXSynID has some limitations. Attributes such as age and subtle facial features are not always represented reliably, and similarity-based filtering may unintentionally skew demographic distributions. In contrast to ONOT~\cite{onot}, which enforces strict ICAO compliance~\cite{icao} at the cost of a 73\% reduction in dataset size, FLUXSynID does not perform explicit ICAO filtering. While document-style images are generated to generally align with ICAO standards (e.g., frontal pose, neutral expression), we prioritize dataset size and embedding space diversity over strict adherence, enhancing flexibility for a broader range of biometric research applications.

\section{Acknowledgments}
This research was funded by the European Union under the Horizon Europe programme, Grant Agreement No. 101121280. Views and opinions expressed are however those of the author(s) only and do not necessarily reflect the views of the EU/Executive Agency. Neither the EU nor the granting authority can be held responsible for them.

{
    \small
    \bibliographystyle{ieeenat_fullname}
    \bibliography{main}
}

\end{document}


%% file: main.bbl
\begin{thebibliography}{55}
\providecommand{\natexlab}[1]{#1}
\providecommand{\url}[1]{\texttt{#1}}
\expandafter\ifx\csname urlstyle\endcsname\relax
  \providecommand{\doi}[1]{doi: #1}\else
  \providecommand{\doi}{doi: \begingroup \urlstyle{rm}\Url}\fi

\bibitem[Bae et~al.(2022)Bae, de~La~Gorce, Baltru\v{s}aitis, Hewitt, Chen, Valentin, Cipolla, and Shen]{Bae2022DigiFace1M1M}
Gwangbin Bae, Martin de La~Gorce, Tadas Baltru\v{s}aitis, Charlie Hewitt, Dong Chen, Julien Valentin, Roberto Cipolla, and JingJing Shen.
\newblock Digiface-1m: 1 million digital face images for face recognition.
\newblock \emph{2023 IEEE/CVF Winter Conference on Applications of Computer Vision (WACV)}, pages 3515--3524, 2022.

\bibitem[Banerjee and Ross(2021)]{Banerjee2021ConditionalID}
Sudipta Banerjee and Arun Ross.
\newblock Conditional identity disentanglement for differential face morph detection.
\newblock \emph{2021 IEEE International Joint Conference on Biometrics (IJCB)}, pages 1--8, 2021.

\bibitem[{Black Forest Labs}(2024)]{flux}
{Black Forest Labs}.
\newblock {FLUX}.
\newblock \url{https://github.com/black-forest-labs/flux}, 2024.
\newblock Accessed: 2025-05-02.

\bibitem[Boutros et~al.(2022{\natexlab{a}})Boutros, Huber, Siebke, Rieber, and Damer]{Boutros2022SFacePA}
Fadi Boutros, Marco Huber, Patrick Siebke, Tim Rieber, and Naser Damer.
\newblock Sface: Privacy-friendly and accurate face recognition using synthetic data.
\newblock \emph{2022 IEEE International Joint Conference on Biometrics (IJCB)}, pages 1--11, 2022{\natexlab{a}}.

\bibitem[Boutros et~al.(2022{\natexlab{b}})Boutros, Klemt, Fang, Kuijper, and Damer]{Boutros2022UnsupervisedFR}
Fadi Boutros, Marcel Klemt, Meiling Fang, Arjan Kuijper, and Naser Damer.
\newblock Unsupervised face recognition using unlabeled synthetic data.
\newblock \emph{2023 IEEE 17th International Conference on Automatic Face and Gesture Recognition (FG)}, pages 1--8, 2022{\natexlab{b}}.

\bibitem[Boutros et~al.(2023)Boutros, Grebe, Kuijper, and Damer]{Boutros2023IDiffFaceSF}
Fadi Boutros, Jonas~Henry Grebe, Arjan Kuijper, and Naser Damer.
\newblock Idiff-face: Synthetic-based face recognition through fizzy identity-conditioned diffusion models.
\newblock \emph{2023 IEEE/CVF International Conference on Computer Vision (ICCV)}, pages 19593--19604, 2023.

\bibitem[Chaudhary et~al.(2021)Chaudhary, Aghdaie, Soleymani, Dawson, and Nasrabadi]{Chaudhary2021DifferentialMF}
Baaria Chaudhary, Poorya Aghdaie, Sobhan Soleymani, Jeremy Dawson, and Nasser Nasrabadi.
\newblock Differential morph face detection using discriminative wavelet sub-bands.
\newblock \emph{2021 IEEE/CVF Conference on Computer Vision and Pattern Recognition Workshops (CVPRW)}, pages 1425--1434, 2021.

\bibitem[Chen et~al.(2024)Chen, Wang, Cao, Liu, Gao, Cui, et~al.]{InternVL2_5}
Zhe Chen, Weiyun Wang, Yue Cao, Yangzhou Liu, Zhangwei Gao, Erfei Cui, et~al.
\newblock Expanding performance boundaries of open-source multimodal models with model, data, and test-time scaling.
\newblock \emph{ArXiv}, abs/2412.05271, 2024.

\bibitem[cocktailpeanut(2025)]{fluxgym}
cocktailpeanut.
\newblock {FluxGym}.
\newblock \url{https://github.com/cocktailpeanut/fluxgym}, 2025.
\newblock Accessed: 2025-05-02.

\bibitem[comfyanonymous(2023)]{comfyui}
comfyanonymous.
\newblock Comfyui: The most powerful and modular diffusion model gui, api, and backend with a graph/nodes interface.
\newblock \url{https://github.com/comfyanonymous/ComfyUI}, 2023.
\newblock Accessed: 2025-05-02.

\bibitem[DeBruine and Jones(2017)]{frll}
Lisa DeBruine and Benedict Jones.
\newblock Face research lab london set.
\newblock \url{https://figshare.com/articles/dataset/Face_Research_Lab_London_Set/5047666/3}, 2017.

\bibitem[Deng et~al.(2019)Deng, Guo, Xue, and Zafeiriou]{arcface}
Jiankang Deng, Jia Guo, Niannan Xue, and Stefanos Zafeiriou.
\newblock Arcface: Additive angular margin loss for deep face recognition.
\newblock In \emph{Proceedings of the IEEE/CVF Conference on Computer Vision and Pattern Recognition (CVPR)}, 2019.

\bibitem[Dess{\`i} et~al.(2023)Dess{\`i}, Bevilacqua, Gualdoni, Rakotonirina, Franzon, and Baroni]{Dess2023CrossDomainIC}
Roberto Dess{\`i}, Michele Bevilacqua, Eleonora Gualdoni, Nathana{\"e}l~Carraz Rakotonirina, Francesca Franzon, and Marco Baroni.
\newblock Cross-domain image captioning with discriminative finetuning.
\newblock \emph{2023 IEEE/CVF Conference on Computer Vision and Pattern Recognition (CVPR)}, pages 6935--6944, 2023.

\bibitem[Domenico et~al.(2024)Domenico, Borghi, Franco, and Maltoni]{onot}
Nicol{\`o}~Di Domenico, Guido Borghi, Annalisa Franco, and Davide Maltoni.
\newblock Onot: a high-quality icao-compliant synthetic mugshot dataset.
\newblock \emph{2024 IEEE 18th International Conference on Automatic Face and Gesture Recognition (FG)}, pages 1--10, 2024.

\bibitem[Elsheikh et~al.(2024)Elsheikh, Mohamed, Abou-Taleb, and Ata]{Elsheikh2024ImprovedFE}
Reham~A. Elsheikh, M.~A. Mohamed, Ahmed~Mohamed Abou-Taleb, and Mohamed~Maher Ata.
\newblock Improved facial emotion recognition model based on a novel deep convolutional structure.
\newblock \emph{Scientific Reports}, 14, 2024.

\bibitem[Esser et~al.(2024)Esser, Kulal, Blattmann, Entezari, M\"{u}ller, Saini, Levi, Lorenz, Sauer, Boesel, Podell, Dockhorn, English, and Rombach]{rectifying_flow_transformer}
Patrick Esser, Sumith Kulal, Andreas Blattmann, Rahim Entezari, Jonas M\"{u}ller, Harry Saini, Yam Levi, Dominik Lorenz, Axel Sauer, Frederic Boesel, Dustin Podell, Tim Dockhorn, Zion English, and Robin Rombach.
\newblock Scaling rectified flow transformers for high-resolution image synthesis.
\newblock In \emph{Proceedings of the 41st International Conference on Machine Learning}, pages 12606--12633. PMLR, 2024.

\bibitem[Fisch et~al.(2020)Fisch, Lee, Chang, Clark, and Barzilay]{Fisch2020CapWAPIC}
Adam Fisch, Kenton Lee, Ming-Wei Chang, Jonathan Clark, and Regina Barzilay.
\newblock Capwap: Image captioning with a purpose.
\newblock In \emph{Conference on Empirical Methods in Natural Language Processing}, 2020.

\bibitem[Frontex(2012)]{frontex}
Frontex.
\newblock \emph{Best practice operational guidelines for Automated Border Control (ABC) systems – Research and development unit}.
\newblock Publications Office of the European Union, Warsaw, Poland, 2012.

\bibitem[Gao et~al.(2025)Gao, Lu, Walters, Zhou, Chu, Zhang, Zhang, Jia, Zhao, Fan, and Zhang]{eraseanything}
Daiheng Gao, Shilin Lu, Shaw Walters, Wenbo Zhou, Jiaming Chu, Jie Zhang, Bang Zhang, Mengxi Jia, Jian Zhao, Zhaoxin Fan, and Weiming Zhang.
\newblock Eraseanything: Enabling concept erasure in rectified flow transformers, 2025.

\bibitem[Guo et~al.(2025)Guo, Zhang, Liu, Zhong, Zhang, Wan, and Zhang]{live_portrait}
Jianzhu Guo, Dingyun Zhang, Xiaoqiang Liu, Zhizhou Zhong, Yuan Zhang, Pengfei Wan, and Di Zhang.
\newblock Liveportrait: Efficient portrait animation with stitching and retargeting control, 2025.

\bibitem[Guo et~al.(2024)Guo, Wu, Chen, Chen, Zhang, and He]{pulid}
Zinan Guo, Yanze Wu, Zhuowei Chen, Lang Chen, Peng Zhang, and Qian He.
\newblock Pulid: Pure and lightning id customization via contrastive alignment.
\newblock In \emph{Advances in Neural Information Processing Systems}, 2024.

\bibitem[Ho(2022)]{cfg}
Jonathan Ho.
\newblock Classifier-free diffusion guidance.
\newblock \emph{ArXiv}, abs/2207.12598, 2022.

\bibitem[Hu et~al.(2021)Hu, Shen, Wallis, Allen-Zhu, Li, Wang, Wang, and Chen]{lora}
Edward Hu, Yelong Shen, Phillip Wallis, Zeyuan Allen-Zhu, Yuanzhi Li, Shean Wang, Lu Wang, and Weizhu Chen.
\newblock Lora: Low-rank adaptation of large language models, 2021.

\bibitem[Huber et~al.(2023)Huber, Luu, Boutros, Kuijper, and Damer]{Huber2023BiasAD}
Marco Huber, An Luu, Fadi Boutros, Arjan Kuijper, and Naser Damer.
\newblock Bias and diversity in synthetic-based face recognition.
\newblock \emph{2024 IEEE/CVF Winter Conference on Applications of Computer Vision (WACV)}, pages 6203--6214, 2023.

\bibitem[{ISO/IEC JTC1 SC17 WG3}(2018)]{icao}
{ISO/IEC JTC1 SC17 WG3}.
\newblock Portrait quality (reference facial images for mrtd), version 1.0.
\newblock Technical report, International Civil Aviation Organization, 2018.

\bibitem[Karras et~al.(2020)Karras, Laine, Aittala, Hellsten, Lehtinen, and Aila]{stylegan2}
Tero Karras, Samuli Laine, Miika Aittala, Janne Hellsten, Jaakko Lehtinen, and Timo Aila.
\newblock Analyzing and improving the image quality of stylegan.
\newblock In \emph{2020 IEEE/CVF Conference on Computer Vision and Pattern Recognition (CVPR)}, pages 8107--8116, 2020.

\bibitem[Karras et~al.(2021)Karras, Aittala, Laine, H{\"a}rk{\"o}nen, Hellsten, Lehtinen, and Aila]{Karras2021AliasFreeGA}
Tero Karras, Miika Aittala, Samuli Laine, Erik H{\"a}rk{\"o}nen, Janne Hellsten, Jaakko Lehtinen, and Timo Aila.
\newblock Alias-free generative adversarial networks.
\newblock In \emph{Neural Information Processing Systems}, 2021.

\bibitem[Kim et~al.(2022)Kim, Jain, and Liu]{adaface}
Minchul Kim, Anil~Kumar Jain, and Xiaoming Liu.
\newblock Adaface: Quality adaptive margin for face recognition.
\newblock In \emph{Proceedings of the IEEE/CVF Conference on Computer Vision and Pattern Recognition (CVPR)}, pages 18750--18759, 2022.

\bibitem[Kim et~al.(2023)Kim, Liu, Jain, and Liu]{Kim2023DCFaceSF}
Minchul Kim, Feng Liu, Anil Jain, and Xiaoming Liu.
\newblock Dcface: Synthetic face generation with dual condition diffusion model.
\newblock \emph{2023 IEEE/CVF Conference on Computer Vision and Pattern Recognition (CVPR)}, pages 12715--12725, 2023.

\bibitem[Kreiss et~al.(2021)Kreiss, Fang, Goodman, and Potts]{Kreiss2021ConcadiaTI}
Elisa Kreiss, Fei Fang, Noah Goodman, and Christopher Potts.
\newblock Concadia: Towards image-based text generation with a purpose.
\newblock In \emph{Conference on Empirical Methods in Natural Language Processing}, 2021.

\bibitem[Lakshmi et~al.(2020)Lakshmi, Wittenbrink, Correll, and Ma]{chicago_db_3}
Anjana Lakshmi, Bernd Wittenbrink, Joshua Correll, and Debbie~S. Ma.
\newblock The india face set: International and cultural boundaries impact face impressions and perceptions of category membership.
\newblock \emph{Frontiers in Psychology}, 12:\penalty0 161, 2020.

\bibitem[Liu et~al.(2024)Liu, Ferrara, Franco, Borghi, and Zhong]{Liu2024DifferentialMA}
Chengcheng Liu, Matteo Ferrara, Annalisa Franco, Guido Borghi, and Dexing Zhong.
\newblock Differential morphing attack detection via triplet-based metric learning and artifact extraction.
\newblock \emph{2024 International Conference of the Biometrics Special Interest Group (BIOSIG)}, pages 1--7, 2024.

\bibitem[Long et~al.(2024)Long, Yao, Zhang, and Peng]{10415238}
Min Long, Quantao Yao, Le-Bing Zhang, and Fei Peng.
\newblock Face de-morphing based on diffusion autoencoders.
\newblock \emph{IEEE Transactions on Information Forensics and Security}, 19:\penalty0 3051--3063, 2024.

\bibitem[Lu et~al.(2023)Lu, Zhou, Bao, Chen, Li, and Zhu]{dpmpp}
Cheng Lu, Yuhao Zhou, Fan Bao, Jianfei Chen, Chongxuan Li, and Jun Zhu.
\newblock Dpm-solver++: Fast solver for guided sampling of diffusion probabilistic models, 2023.

\bibitem[Ma et~al.(2015)Ma, Correll, and Wittenbrink]{chicago_db_1}
Debbie~S. Ma, Joshua Correll, and Bernd Wittenbrink.
\newblock The chicago face database: A free stimulus set of faces and norming data.
\newblock \emph{Behavior Research Methods}, 47:\penalty0 1122--1135, 2015.

\bibitem[Ma et~al.(2020)Ma, Kantner, and Wittenbrink]{chicago_db_2}
Debbie~S. Ma, Justin Kantner, and Bernd Wittenbrink.
\newblock Chicago face database: Multiracial expansion.
\newblock \emph{Behavior Research Methods}, 2020.

\bibitem[Melzi et~al.(2023)Melzi, Rathgeb, Tolosana, Vera-Rodr{\'i}guez, Lawatsch, Domin, and Schaubert]{gandiffface}
Pietro Melzi, Christian Rathgeb, Rub{\'e}n Tolosana, Rub{\'e}n Vera-Rodr{\'i}guez, Dominik Lawatsch, Florian Domin, and Maxim Schaubert.
\newblock Gandiffface: Controllable generation of synthetic datasets for face recognition with realistic variations.
\newblock \emph{2023 IEEE/CVF International Conference on Computer Vision Workshops (ICCVW)}, pages 3078--3087, 2023.

\bibitem[Melzi et~al.(2024)Melzi, Tolosana, Vera-Rodr{\'i}guez, Kim, Rathgeb, Liu, et~al.]{Melzi2024FRCSynonGoingBA}
Pietro Melzi, Rub{\'e}n Tolosana, Rub{\'e}n Vera-Rodr{\'i}guez, Minchul Kim, Christian Rathgeb, Xiaoming Liu, et~al.
\newblock Frcsyn-ongoing: Benchmarking and comprehensive evaluation of real and synthetic data to improve face recognition systems.
\newblock \emph{Inf. Fusion}, 107:\penalty0 102322, 2024.

\bibitem[Novikov et~al.(2025)Novikov, Vranka, David, and Voronin]{flux_age_estimation}
Alexey~A. Novikov, Miroslav Vranka, Franccois David, and Artem Voronin.
\newblock Can text-to-image generative models accurately depict age? a comparative study on synthetic portrait generation and age estimation.
\newblock \emph{ArXiv}, abs/2502.03420, 2025.

\bibitem[Papantoniou et~al.(2024)Papantoniou, Lattas, Moschoglou, Deng, Kainz, and Zafeiriou]{arc2face}
Foivos~Paraperas Papantoniou, Alexandros Lattas, Stylianos Moschoglou, Jiankang Deng, Bernhard Kainz, and Stefanos Zafeiriou.
\newblock Arc2face: A foundation model for id-consistent human faces.
\newblock In \emph{European Conference on Computer Vision}, 2024.

\bibitem[Phillips et~al.(2005)Phillips, Flynn, Scruggs, Bowyer, Chang, Hoffman, Marques, Min, and Worek]{frgc}
P.~Jonathon Phillips, Patrick~J. Flynn, Todd Scruggs, Kevin~W. Bowyer, Jin Chang, Kevin Hoffman, Joe Marques, Jaesik Min, and William Worek.
\newblock Overview of the face recognition grand challenge.
\newblock In \emph{2005 IEEE Computer Society Conference on Computer Vision and Pattern Recognition (CVPR'05)}, pages 947--954 vol. 1, 2005.

\bibitem[Qin et~al.(2024)Qin, Wang, Deng, Wang, Chen, Hu, and Deng]{10216308}
Lixiong Qin, Mei Wang, Chao Deng, Ke Wang, Xi Chen, Jiani Hu, and Weihong Deng.
\newblock Swinface: A multi-task transformer for face recognition, expression recognition, age estimation and attribute estimation.
\newblock \emph{IEEE Transactions on Circuits and Systems for Video Technology}, 34\penalty0 (4):\penalty0 2223--2234, 2024.

\bibitem[Qiu et~al.(2021)Qiu, Yu, Gong, Li, Liu, and Tao]{Qiu2021SynFaceFR}
Haibo Qiu, Baosheng Yu, Dihong Gong, Zhifeng Li, Wei Liu, and Dacheng Tao.
\newblock Synface: Face recognition with synthetic data.
\newblock \emph{2021 IEEE/CVF International Conference on Computer Vision (ICCV)}, pages 10860--10870, 2021.

\bibitem[Radford et~al.(2021)Radford, Kim, Hallacy, Ramesh, Goh, Agarwal, Sastry, Askell, Mishkin, Clark, Krueger, and Sutskever]{clip}
Alec Radford, Jong~Wook Kim, Chris Hallacy, Aditya Ramesh, Gabriel Goh, Sandhini Agarwal, Girish Sastry, Amanda Askell, Pamela Mishkin, Jack Clark, Gretchen Krueger, and Ilya Sutskever.
\newblock Learning transferable visual models from natural language supervision.
\newblock In \emph{Proceedings of the 38th International Conference on Machine Learning}, pages 8748--8763. PMLR, 2021.

\bibitem[Raffel et~al.(2020)Raffel, Shazeer, Roberts, Lee, Narang, Matena, Zhou, Li, and Liu]{t5}
Colin Raffel, Noam Shazeer, Adam Roberts, Katherine Lee, Sharan Narang, Michael Matena, Yanqi Zhou, Wei Li, and Peter Liu.
\newblock Exploring the limits of transfer learning with a unified text-to-text transformer.
\newblock \emph{Journal of Machine Learning Research}, 21\penalty0 (140):\penalty0 1--67, 2020.

\bibitem[Richardson et~al.(2021)Richardson, Alaluf, Patashnik, Nitzan, Azar, Shapiro, and Cohen-Or]{psp}
Elad Richardson, Yuval Alaluf, Or Patashnik, Yotam Nitzan, Yaniv Azar, Stav Shapiro, and Daniel Cohen-Or.
\newblock Encoding in style: a stylegan encoder for image-to-image translation.
\newblock In \emph{2021 IEEE/CVF Conference on Computer Vision and Pattern Recognition (CVPR)}, pages 2287--2296, 2021.

\bibitem[Robbins et~al.(2024)Robbins, Bertocco, and Boult]{Robbins2024DaliIDDL}
Wes Robbins, Gabriel Bertocco, and Terrance~E. Boult.
\newblock Daliid: Distortion-adaptive learned invariance for identification—a robust technique for face recognition and person re-identification.
\newblock \emph{IEEE Access}, 12:\penalty0 55784--55799, 2024.

\bibitem[Rombach et~al.(2022)Rombach, Blattmann, Lorenz, Esser, and Ommer]{latent_diffusion}
Robin Rombach, Andreas Blattmann, Dominik Lorenz, Patrick Esser, and Bj\"orn Ommer.
\newblock High-resolution image synthesis with latent diffusion models.
\newblock In \emph{Proceedings of the IEEE/CVF Conference on Computer Vision and Pattern Recognition (CVPR)}, pages 10684--10695, 2022.

\bibitem[Roy et~al.(2024)Roy, Kathania, Sharma, Dey, and Ansari]{Roy2024ResEmoteNetBA}
Arnab~Kumar Roy, Hemant~Kumar Kathania, Adhitiya Sharma, Abhishek Dey, and Md. Sarfaraj~Alam Ansari.
\newblock Resemotenet: Bridging accuracy and loss reduction in facial emotion recognition.
\newblock \emph{IEEE Signal Processing Letters}, 32:\penalty0 491--495, 2024.

\bibitem[Ruiz et~al.(2022)Ruiz, Li, Jampani, Pritch, Rubinstein, and Aberman]{dreambooth}
Nataniel Ruiz, Yuanzhen Li, Varun Jampani, Yael Pritch, Michael Rubinstein, and Kfir Aberman.
\newblock Dreambooth: Fine tuning text-to-image diffusion models for subject-driven generation.
\newblock \emph{2023 IEEE/CVF Conference on Computer Vision and Pattern Recognition (CVPR)}, pages 22500--22510, 2022.

\bibitem[Shukla and Ross(2024)]{Shukla2024FacialDV}
Nitish Shukla and Arun Ross.
\newblock Facial demorphing via identity preserving image decomposition.
\newblock \emph{2024 IEEE International Joint Conference on Biometrics (IJCB)}, pages 1--10, 2024.

\bibitem[ToTheBeginning(2024)]{pulid_github}
ToTheBeginning.
\newblock Pulid for flux.
\newblock \url{https://github.com/ToTheBeginning/PuLID/blob/main/docs/pulid_for_flux.md}, 2024.
\newblock Accessed: 2025-05-02.

\bibitem[van~der Maaten and Hinton(2008)]{tsne}
Laurens van~der Maaten and Geoffrey Hinton.
\newblock Visualizing data using t-sne.
\newblock \emph{Journal of Machine Learning Research}, 9\penalty0 (86):\penalty0 2579--2605, 2008.

\bibitem[Venkatesh et~al.(2021)Venkatesh, Ramachandra, Raja, and Busch]{face_morphing_attacks}
Sushma Venkatesh, Raghavendra Ramachandra, Kiran Raja, and Christoph Busch.
\newblock Face morphing attack generation and detection: A comprehensive survey.
\newblock \emph{IEEE Transactions on Technology and Society}, 2\penalty0 (3):\penalty0 128--145, 2021.

\bibitem[Yang et~al.(2024)Yang, Yang, Zhang, Hui, Zheng, Yu, et~al.]{qwen2.5}
An Yang, Baosong Yang, Beichen Zhang, Binyuan Hui, Bo Zheng, Bowen Yu, et~al.
\newblock Qwen2.5 technical report.
\newblock \emph{arXiv preprint arXiv:2412.15115}, 2024.

\end{thebibliography}
